\documentclass{article}

\PassOptionsToPackage{numbers, compress, sort}{natbib}
\usepackage[preprint]{neurips_2026}

\usepackage[utf8]{inputenc} % allow utf-8 input
\usepackage[T1]{fontenc}    % use 8-bit T1 fonts
\usepackage{hyperref}       % hyperlinks
\usepackage{url}            % simple URL typesetting
\usepackage{booktabs}       % professional-quality tables
\usepackage{amsfonts}       % blackboard math symbols
\usepackage{nicefrac}       % compact symbols for 1/2, etc.
\usepackage{microtype}      % microtypography
\usepackage{xcolor}         % colors
\usepackage{amsmath, amssymb}
\usepackage{enumitem}
\usepackage{algorithm}
\usepackage{algorithmic}
\usepackage{array}
\usepackage{dsfont}
\usepackage{booktabs}
\usepackage{multirow}
\usepackage{xcolor}
\usepackage{graphicx, subcaption}
\usepackage{wrapfig}
\usepackage{tabularx}
\usepackage{makecell}
\usepackage{hyperref}
\usepackage[table]{xcolor}
\newcommand{\up}[1]{\textcolor{teal!70!black}{\small$_{+#1}$}}

\newcommand{\repro}{$^{\dagger}$}

\def\ie{\emph{i.e.}}

\def\vs{\emph{vs. }}
\def\wrt{\emph{w.r.t. }}

\title{What Happens Before Decoding? Prefill Determines GUI Grounding in VLMs}

\author{
Jiaping Lin$^{1}$ \quad
Fei Shen$^{2}$ \quad
Junzhe Li$^{3}$ \quad
Ping Nie$^{4}$ \quad
Fei Yu$^{1}$ \quad
Ming Li$^{1}$\thanks{Corresponding author.} \quad
Haizhou Li$^{5}$ \\[0.5em]
$^{1}$Guangming Laboratory \quad $^{2}$National University of Singapore \\
$^{3}$Peking University \quad $^{4}$University of Waterloo \\
$^{5}$The Chinese University of Hong Kong (Shenzhen) \\
}

\begin{document}

\maketitle

\begin{abstract}
Existing training-free approaches for GUI grounding often rely on multiple inference runs, such as iterative cropping or candidate aggregation, to identify target elements.
Despite this additional computation, each forward pass still independently interprets the instruction and parses the visual layout, without enabling progressive interaction among visual tokens.
In this paper, we study what happens during GUI grounding in Vision-Language Models (VLMs) and identify a previously overlooked bottleneck.
We show that grounding follows a two-stage paradigm: the \emph{prefill stage} determines candidate UI elements, while the \emph{decoding stage} subsequently refines the final coordinates.
This asymmetry establishes prefill as the critical step, as errors in candidate selection cannot be effectively corrected during decoding.
Based on this observation, we propose \textbf{Re-Prefill}, a training-free method that revisits inference by introducing an attention-guided second prefill stage to refine target selection. 
Specifically, visual tokens that consistently receive high attention from the query position, \ie, the final token, across layers are extracted as a preliminary target hypothesis and appended to the input, together with the instruction hidden states, enabling the model to deeply re-think its decision before coordinate generation.
Experiments across four VLMs and five benchmarks, including ScreenSpot-Pro, ScreenSpot-V2, OSWorld-G, UI-Vision, and MMBench-GUI, demonstrate consistent improvements without additional training, with gains of up to 4.3\% on ScreenSpot-Pro. Code will be available at 
\url{https://github.com/linjiaping1/Re-Prefill}.
\end{abstract}

\section{Introduction}
\label{sec:intro}

GUI grounding aims to predict the coordinates of a target UI element from a screenshot guided by a natural language instruction, and is a fundamental capability for autonomous agents~\cite{survey_nguyen2025gui, survey_zhang2024large, survey_wang2024gui}.
Vision-Language Models (VLMs) have become the dominant approach to this task, achieving strong performance through large-scale supervised and reinforcement training~\citep{gta1, guig2, maiui, guiowl}.
However, such performance relies on extensive annotation and substantial computing, leading to high labor and computational costs~\cite{cogagent, osatlas, uground}.

To reduce these costs, recent work has explored training-free methods that enhance grounding at inference time~\cite{reguide, regionfocus}.
These approaches typically rely on multiple inference runs, such as iterative cropping to refine visual resolution~\citep{regionfocus, zoomclick, chainofground} or aggregating predictions across runs~\citep{uizoomer, mvp, dimogui}. For example, ZoomClick~\citep{zoomclick} characterizes key properties of zooming and performs iterative crop-and-zoom to increase the effective resolution around targets. CoG~\citep{chainofground} uses context to refine iterative cropping, thereby improving grounding precision. MVP~\citep{mvp} generates predictions from multiple cropped views and aggregates them to achieve more stable grounding accuracy.
While effective, they incur substantial computational overhead.
More importantly, our analysis reveals that stacking multiple runs does not enable interaction between visual tokens across runs, and therefore cannot correct errors introduced in earlier passes.

\begin{figure}[t]
  \centering

  \begin{subfigure}[b]{0.36\textwidth}
    \centering
    \includegraphics[width=\linewidth]{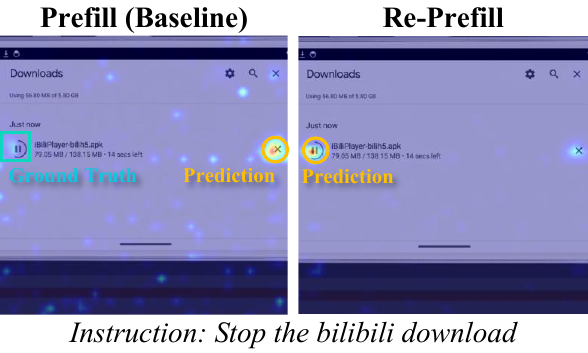}
    \caption{}
    \label{fig:analysis_heatmap}
  \end{subfigure}
  \hspace{0.01\textwidth}
  \begin{subfigure}[b]{0.29\textwidth}
    \centering
    \includegraphics[width=\linewidth]{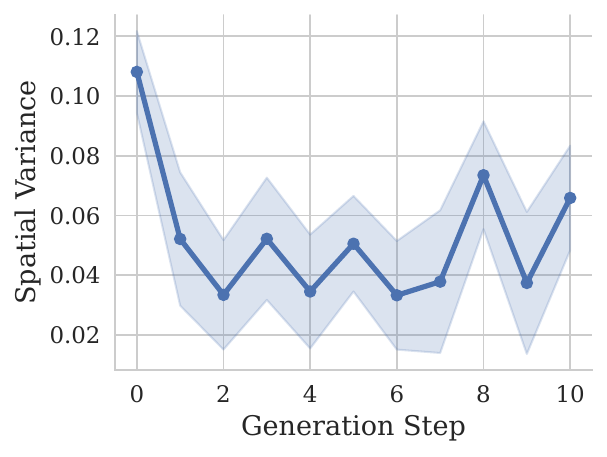}
    \caption{}
    \label{fig:analysis_spatial}
  \end{subfigure}
  \hspace{0.01\textwidth}
  \begin{subfigure}[b]{0.29\textwidth}
    \centering
    \includegraphics[width=\linewidth]{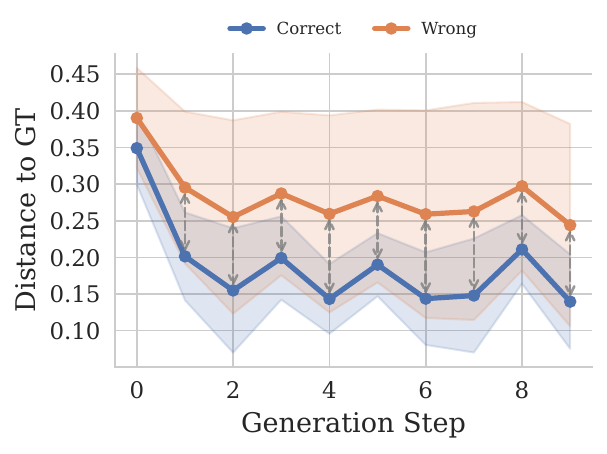}
    \caption{}
    \label{fig:analysis_error}
  \end{subfigure}

  \caption{
  \textbf{Prefill \vs Re-Prefill \vs Decoding.}
  \emph{(a)} Query-position attention heatmaps over visual tokens. Re-Prefill produces a sharper, more focused distribution that disambiguates the correct target from other candidates. Additional visualizations are provided in Appendix~\ref{app:attn_visualization}.
  \emph{(b)} Spatial variance of query-position attention across generation steps. The sharp drop after the first generated token shows that target selection is largely completed during the prefill stage.
  \emph{(c)} Deviation of the attention centroid from the ground-truth, separated into correct (blue) and incorrect (orange) predictions. The persistent gap in incorrect cases indicates that prefill errors cannot be corrected during decoding. Extended analysis across multiple models and benchmarks is provided in Appendix~\ref{app:cross_bench}.
}
\label{fig:analysis}
\vspace{-2em}
\end{figure}

In this work, we attempt to address a more fundamental question: \emph{where do grounding errors originate within inference?} We investigate this question by asking: \emph{what happens before decoding?} We perform a thorough experimental analysis by tracing the query-position attention, \ie the attention from the final token to the visual tokens, at each generation step. We then compute both its spatial variance, reflecting the concentration of attention over image regions, and the deviation of its centroid from the ground-truth target. 
In Figure.~\ref{fig:analysis}, we show the visual-token attention distribution from the query position at the prefill stage and the spatial variance of attention, the deviation of the attention centroid from the ground truth \wrt the decoding step.
Step $t=0$ on the horizontal axis corresponds to the prefill stage, while $t=1$ indicates the start of decoding.
The results show that during prefill, attention is scattered on a couple of candidate regions, coarsely determining the target element scope.
During decoding, coordinates are generated autoregressively by concentrating on one specific region, which can be demonstrated by attention variance decrease.
Importantly, the deviation from the ground truth remains large for incorrect predictions, indicating that errors introduced during prefill persist throughout decoding.

These observations establish the prefill stage as the decisive step for grounding, where accurate target selection is critical for subsequent decoding.
Based on this insight, we propose \textbf{Re-Prefill}, a training-free method that revisits inference by introducing an attention-guided second prefill stage to refine attention concentration and improve target selection.
Specifically, after the initial prefill computation, we identify visual tokens that consistently receive high attention from the query position across layers, which represent the model's preliminary target hypothesis.
These tokens, as well as the encoded instruction hidden states, are appended to the original input for re-prefill computation, enabling the model to re-evaluate its target selection before coordinate generation. The illustration in Figure~\ref{fig:analysis} illustrates that Re-Prefill improves target selection by enabling the model to focus on the correct UI element before decoding.

We evaluate Re-Prefill on four VLMs of varying scales~\citep{maiui, guiowl, qwen3vl} across five GUI grounding benchmarks, including ScreenSpot-Pro~\citep{sspro}, ScreenSpot-V2~\citep{osatlas}, OSWorld-G~\citep{osworldg}, UI-Vision~\citep{uivision}, and MMBench-GUI~\citep{mmbench}.
Re-Prefill consistently improves grounding accuracy across all models and benchmarks without additional training, achieving gains of up to 4.3\% on ScreenSpot-Pro and 2.7\% on OSWorld-G.

Our contributions in this work are summarized as follows:
\begin{itemize}[leftmargin=2em, itemsep=2pt]
\item We identify a two-stage inference paradigm in GUI grounding, where the prefill stage determines target selection and the decoding stage generates coordinates conditioned on it.
\item We propose Re-Prefill, a training-free method that improves target selection by revisiting inference with an attention-guided second prefill stage, orthogonal to existing GUI grounding methods.
\item We demonstrate consistent improvements across models and benchmarks, establishing prefill as a key bottleneck and providing a new perspective for studying GUI grounding.
\end{itemize}

\section{Revisiting GUI Grounding Inference}
\label{sec:analysis}

\subsection{Preliminaries}
\label{sec:preliminary}

GUI grounding aims to predict the coordinates $(x, y)$ of a target UI element given a screenshot $I$ and a natural language instruction $q$. 
A decoder-only VLM processes the input as a sequence of tokens composed of three segments: system prompt tokens $\mathbf{S}$, visual tokens $\mathbf{V}$, and instruction tokens $\mathbf{T}$. 
The concatenated input is denoted as $\mathbf{x} = [\mathbf{S}; \mathbf{V}; \mathbf{T}]$.

Inference consists of two stages. 
In the \emph{prefill stage}, the full sequence $\mathbf{x}$ is processed through $L$ decoder layers in a single forward pass, where self-attention is applied over all tokens and the resulting key--value pairs are stored in a KV cache $\mathcal{C}$. 
We denote this process as $f_{\mathrm{prefill}}(\mathbf{x}) \rightarrow (\mathbf{h}_q, \mathcal{C})$, where $\mathbf{h}_q$ is the hidden state at the final token position (the query position), which is used to initiate coordinate prediction.
In the subsequent \emph{decoding stage}, coordinate tokens are generated autoregressively, with each step conditioned on the KV cache and previously generated tokens: $y_t = f_{\mathrm{decode}}(y_{<t},\, \mathcal{C})$.

\subsection{Prefill as the Bottleneck of GUI Grounding}
\label{sec:bottleneck}
To better understand where grounding errors originate, we analyze the attention dynamics of VLMs during inference. In particular, we investigate two key questions: (1) when target region selection is determined, and (2) whether an incorrect selection can be corrected during decoding.

We employ a unified probing mechanism based on query-position attention. Let $\mathbf{a}_t \in \mathbb{R}^{N_v}$ denote the attention from the query position to visual tokens at generation step $t$, where $t=0$ corresponds to the prefill stage and $t \ge 1$ to decoding steps. To reduce layer-wise variation, $\mathbf{a}_t$ is averaged across all decoder layers. All results are reported on ScreenSpot-Pro with Qwen3-VL-8B-Instruct, with extended analysis across multiple models and benchmarks provided in Appendix~\ref{app:cross_bench}.

\noindent{\textbf{Early attention concentration determines target selection. }}
We first examine how attention evolves spatially across generation steps. The spatial concentration of $\mathbf{a}_t$ is measured by its variance,
\begin{equation}
\sigma_t^2
\;=\;
\sum_{i=1}^{N_v} a_t(i)\,\bigl\|\mathbf{p}_i - \boldsymbol{\mu}_t\bigr\|^2,
\quad
\boldsymbol{\mu}_t
\;=\;
\sum_{i=1}^{N_v} a_t(i)\,\mathbf{p}_i,
\end{equation}
where $\mathbf{p}_i$ is the 2-D coordinate of $v_i$ and $\boldsymbol{\mu}_t$ is the attention centroid. A large $\sigma_t^2$ indicates dispersed attention, while a small value indicates concentration on a localized region.
As shown in Figure~\ref{fig:analysis}(b), $\sigma_t^2$ is high at $t=0$, suggesting that attention during prefill is distributed over multiple candidate regions. At $t=1$, the variance drops sharply, indicating that attention rapidly concentrates on a single region once decoding begins. Thereafter, the variance remains low with only minor fluctuations.

This behavior suggests that target region selection is largely completed during the prefill stage, and that decoding primarily refines coordinates within the selected region. The small fluctuations during decoding likely arise from attending to local contextual layout, which helps improve coordinate precision. Similar patterns are observed across multiple benchmarks, indicating that this behavior is consistent across models and datasets.

\noindent{\textbf{Prefill errors persist throughout decoding. }}
We next examine whether incorrect target selection can be corrected during decoding. Samples are divided into Correct and Wrong groups according to whether the predicted coordinates fall inside the ground-truth bounding box. For each group, we measure the normalized distance between the attention centroid $\boldsymbol{\mu}_t$ and the ground-truth center $\mathbf{g}$:
\begin{equation}
d_t \;=\; \frac{\bigl\|\boldsymbol{\mu}_t - \mathbf{g}\bigr\|}{D},
\end{equation}
where $D$ is the image diagonal.
As shown in Figure~\ref{fig:analysis}(c), both groups exhibit a decrease in distance from $t=0$ to $t=1$, consistent with the attention concentration behavior observed earlier. After $t=1$, the trajectories diverge. The Correct group stabilizes near the target, whereas the Wrong group remains significantly farther away, with the gap persisting throughout decoding.
This persistent discrepancy indicates that errors introduced during prefill are rarely corrected during decoding. 
This limitation arises from the autoregressive inference process, where decoding conditions on cached representations without recomputing attention over visual tokens. Consequently, misalignment formed during prefill propagats through all subsequent decoding steps.

These observations identify the prefill stage as a structural bottleneck in GUI grounding, as it determines target selection while its errors remain irreversible. Motivated by this finding, we propose to enable explicit re-evaluation of target selection before decoding, leading to the Re-Prefill method described in Section~\ref{sec:reprefill}.

\section{Re-Prefill}
\label{sec:reprefill}

\begin{figure}[t]
  \centering
  \includegraphics[width=\textwidth]{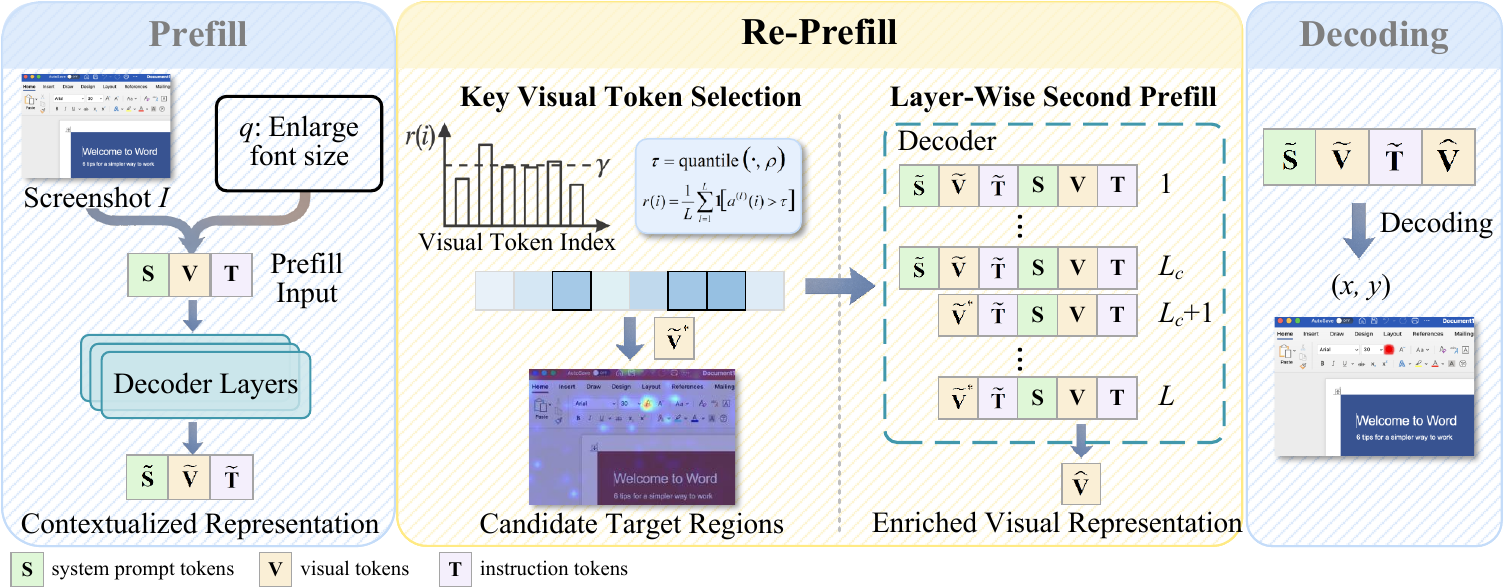}
  \caption{
    \textbf{Overview of Re-Prefill.}
    \emph{(1) Prefill.} The input $[\mathbf{S}; \mathbf{V}; \mathbf{T}]$ is processed through $L$ decoder layers to obtain contextualized representations $[\tilde{\mathbf{S}}; \tilde{\mathbf{V}}; \tilde{\mathbf{T}}]$.
    \emph{(2) Key visual token selection.} Visual tokens that consistently receive high attention across layers are selected as $\tilde{\mathbf{V}}^{*}$, representing candidate target regions.
    \emph{(3) Layer-wise second prefill.} A copy of the original input is re-encoded with layer-wise prefix modification, transitioning from full first-prefill representations to key visual tokens, producing enriched visual representations $\hat{\mathbf{V}}$.
    \emph{(4) Decoding.} Coordinate tokens are generated autoregressively from the combined context $\mathcal{C}^{*} = [\tilde{\mathbf{S}}; \tilde{\mathbf{V}}; \tilde{\mathbf{T}}; \hat{\mathbf{V}}]$.
  }
  \label{fig:overview}
  \vspace{-1.5em}
\end{figure}

% --- Overview ---
Building on the above analysis, we propose \textbf{Re-Prefill}, a training-free method that enhances GUI grounding through an attention-guided second prefill stage. 
Re-Prefill first identifies key visual tokens that encode candidate target regions from the initial prefill, and then performs a second prefill over the original input, where these tokens are incorporated as guidance to refine target region selection before decoding.
An overview of the full pipeline is shown in Figure~\ref{fig:overview}.

% --- Step 1 ---
\noindent{\textbf{Step 1: Initial prefill and attention extraction. }}
A standard prefill stage is first applied to the input $\mathbf{x} = [\mathbf{S}; \mathbf{V}; \mathbf{T}]$, producing contextualized representations $\tilde{\mathbf{S}}$, $\tilde{\mathbf{V}}$, and $\tilde{\mathbf{T}}$, together with the KV cache $\mathcal{C}_1$.
To support subsequent visual token filtering, we additionally extract per-layer attention maps from the query position to visual tokens. Specifically, for each decoder layer $l$, the attention $\mathbf{a}^{(l)} \in \mathbb{R}^{N_v}$ is computed and averaged across heads to capture cross-layer visual relevance signals.

% --- Step 2 ---
\noindent{\textbf{Step 2: Key visual token selection.}}
The initial prefill produces contextualized visual representations $\tilde{\mathbf{V}}$ and query-to-visual attention maps. Based on qualitative visualizations (Section~\ref{sec:main_results} and Appendix~\ref{app:attn_visualization}), we observe that the prefill-stage attention is often distributed across multiple plausible UI regions, while the region corresponding to the final prediction typically receives relatively high attention. This suggests that the first prefill implicitly narrows the search space to a set of candidate target regions before selecting one for subsequent decoding. 
In this process, visual tokens with relatively high attention act as coarse indicators of these candidate regions.
Motivated by this observation, we seek to re-evaluate the candidate regions before decoding, thereby increasing the likelihood of selecting the correct target. This requires identifying a subset of contextualized visual tokens that reliably represent the candidate regions.
To this end, we introduce a cross-layer consistency filter over $L$ decoder layers. We first define a global high-attention threshold $\tau$ as the $\rho$-quantile over all attention values:
\begin{equation}
\tau = \mathrm{quantile}\!\bigl(\{a^{(l)}(i)\}_{l,i}, \rho\bigr).
\label{eq:threshold}
\end{equation}

For each visual token, we compute its cross-layer activation ratio:
\begin{equation}
r(i) = \frac{1}{L} \sum_{l=1}^{L} \mathds{1}\!\bigl[a^{(l)}(i) > \tau\bigr].
\end{equation}

The \emph{key visual tokens} are then defined as:
\begin{equation}
\tilde{\mathbf{V}}^{*} = \{\, \tilde{v}_i \mid r(i) \ge \gamma \,\},
\label{eq:ratio}
\end{equation}
which correspond to tokens that consistently exhibit high attention, encoding candidate target regions.

% --- Step 3 ---
\noindent{\textbf{Step 3: Layer-wise second prefill. }}
To enable refined target selection, we perform a second prefill over a copy of the original uncontextualized input $[\mathbf{S}; \mathbf{V}; \mathbf{T}]$, guided by the representations obtained from the first prefill. 

The key design is to progressively transition from full-context conditioning to focused conditioning on key visual tokens. This is achieved through a layer-wise prefix injection strategy controlled by a continuity hyperparameter $L_c$, which determines the layer at which the prefix is updated.
Specifically, for layers $1$ to $L_c$, we prepend the complete first-prefill representations, resulting in $[\tilde{\mathbf{S}};\tilde{\mathbf{V}};\tilde{\mathbf{T}};\mathbf{S};\mathbf{V};\mathbf{T}]$, so as to preserve global semantic alignment.
For layers $L_c + 1$ to $L$, we replace the full visual representations with the selected key visual tokens and remove $\tilde{\mathbf{S}}$, yielding $[\tilde{\mathbf{V}}^{*};\tilde{\mathbf{T}};\mathbf{S};\mathbf{V};\mathbf{T}]$, so that the model focuses on candidate target regions while retaining instruction context. Since the instruction segment is typically short and contains essential information, no additional selection is applied.
After all $L$ layers, the visual tokens in the uncontextualized input are updated into enriched representations, denoted as $\hat{\mathbf{V}}$, which are subsequently used for decoding.

The hyperparameter $L_c$ controls the trade-off between preserving semantic continuity and enabling effective re-focusing. Its impact is analyzed in Section~\ref{sec:ablation}.

% --- Step 4 ---
\noindent{\textbf{Step 4: Decoding. }}
The final stage produces coordinate tokens through autoregressive decoding. The decoding context $\mathcal{C}^{*}$ is constructed by concatenating the first-prefill representations with the enriched visual tokens $\hat{\mathbf{V}}$ from the second prefill, forming $\mathcal{C}^{*} = [\tilde{\mathbf{S}};\tilde{\mathbf{V}};\tilde{\mathbf{T}};\hat{\mathbf{V}}]$. Coordinate tokens are then generated autoregressively over $\mathcal{C}^{*}$. Pseudocode of the full procedure is provided in Appendix~\ref{app:algorithm}.

\section{Experiments}
\label{sec:experiments}

\subsection{Experimental Setup}
\label{sec:setup}

\noindent{\textbf{Benchmarks.}}
Re-Prefill is evaluated on five benchmarks spanning desktop, mobile, and web interfaces, including ScreenSpot-Pro~\citep{sspro}, ScreenSpot-V2~\citep{osatlas}, OSWorld-G~\citep{osworldg}, UI-Vision~\citep{uivision}, and MMBench-GUI-L2~\citep{mmbench}.

\noindent{\textbf{Base models. }}
Four decoder-only VLMs are considered, including Qwen3-VL-8B-Instruct, Qwen3-VL-32B-Instruct~\citep{qwen3vl}, MAI-UI-8B~\citep{maiui}, and GUI-Owl-1.5-8B-Instruct~\citep{guiowl}.
The two Qwen3-VL models are general-purpose VLMs without GUI-specific post-training, whereas MAI-UI-8B and GUI-Owl-1.5-8B-Instruct are designed specifically for GUI grounding.
This selection covers the primary design configurations in current GUI agents and enables evaluation of whether Re-Prefill generalizes across different base model families.

\noindent{\textbf{Implementation details. }}
Re-Prefill involves three hyperparameters, including the quantile threshold $\rho$, the ratio threshold $\gamma$, and the continuity layer $L_c$.
The values $\rho = 0.8$ and $\gamma = 0.1$ are used throughout, with $L_c = 3$ for the 36-layer 8B models and $L_c = 6$ for the 64-layer 32B model.
On ScreenSpot-Pro, OSWorld-G, UI-Vision, and MMBench-GUI-L2, a lightweight zoom-in strategy is applied to both the baseline and Re-Prefill.
A square region centered on the initial prediction is cropped, upsampled by $2{\times}$, and then reprocessed by the same model to obtain a refined prediction.
The zoom-in strategy is not applied on ScreenSpot-V2, as its low-resolution images already exhibit saturated performance without post-processing.
The results for the base models are reproduced in this work under identical inference settings, whereas results for external baselines are taken from the original papers and are indicated accordingly in each table.
We run 8B models on 8 RTX 4090 GPUs and the 32B model on 4 RTX PRO 6000 GPUs.

\subsection{Main Results}
\label{sec:main_results}

Tables~\ref{tab:main_sspro}--\ref{tab:main_mmbench} report results on the five benchmarks.
The two MVP variants~\citep{mvp} share base models with our evaluation (MVP-8B uses Qwen3-VL-8B-Instruct and MVP-32B uses Qwen3-VL-32B-Instruct), enabling a direct comparison of inference-time strategies under identical underlying VLMs.

\begin{table*}[t]
\centering
\caption{Results on ScreenSpot-Pro. Methods marked with \repro{} are reproduced using the zoom-in strategy adopted in this work.}
\label{tab:main_sspro}
\resizebox{\textwidth}{!}{%
\begin{tabular}{l *{12}{c} >{\columncolor{blue!10}}c}
\toprule
\multirow{2}{*}{\textbf{Model}} &
\multicolumn{2}{c}{\textbf{Development}} &
\multicolumn{2}{c}{\textbf{Creative}} &
\multicolumn{2}{c}{\textbf{CAD}} &
\multicolumn{2}{c}{\textbf{Scientific}} &
\multicolumn{2}{c}{\textbf{Office}} &
\multicolumn{2}{c}{\textbf{OS}} &
\multicolumn{1}{c}{\multirow{2}{*}{\textbf{Overall}}} \\
\cmidrule(lr){2-3} \cmidrule(lr){4-5} \cmidrule(lr){6-7} \cmidrule(lr){8-9} \cmidrule(lr){10-11} \cmidrule(lr){12-13}
 & Text & Icon & Text & Icon & Text & Icon & Text & Icon & Text & Icon & Text & Icon & \cellcolor{white} \\
\midrule
UI-TARS-72B~\citep{uitars} & 63.0 & 17.3 & 57.1 & 15.4 & 18.8 & 12.5 & 64.6 & 20.9 & 63.3 & 26.4 & 42.1 & 15.7 & 38.1 \\
GTA1-32B~\citep{gta1} & 83.1 & 37.9 & 72.2 & 25.9 & 70.1 & 31.3 & 84.7 & 39.1 & 89.3 & 64.2 & 76.6 & 51.7 & 63.6 \\
UI-Ins-32B~\citep{uiins} & 51.8 & 29.7 & 83.1 & 26.9 & 69.7 & 18.9 & 83.3 & 34.5 & 88.7 & 50.9 & 70.1 & 34.8 & 57.0 \\
GuirlVG~\citep{guirlvg} & 64.9 & 7.6 & 42.9 & 11.2 & 28.9 & 9.4 & 63.9 & 16.4 & 63.8 & 26.4 & 43.9 & 13.5 & 36.1 \\
% GUI-Spotlight~\citep{guispotlight} & - & - & - & - & - & - & - & - & - & - & - & - & 52.8 \\
GUI-Cursor-7B~\citep{guicursor} & 80.5 & 33.1 & 65.7 & 18.2 & 62.4 & 25.0 & 83.3 & 32.7 & 84.2 & 43.4 & 65.4 & 31.5 & 56.5 \\
MAI-UI-8B~\citep{maiui} & 78.6 & 58.6 & 78.8 & 46.9 & 80.7 & 43.8 & 86.1 & 49.1 & 88.1 & 81.1 & 76.6 & 51.7 & 70.9 \\
GUI-Owl-1.5-8B-Instruct~\citep{guiowl} & 90.2 & 68.9 & 84.8 & 56.6 & 86.8 & 62.5 & 89.5 & 55.4 & 91.5 & 71.6 & 86.9 & 53.9 & 77.8 \\
\midrule
\multicolumn{14}{l}{\textit{Training-free methods}} \\
RegionFocus~\citep{regionfocus} & 75.3 & 25.5 & 76.3 & 30.8 & 71.6 & 28.1 & 87.5 & 39.1 & 87.0 & 60.4 & 74.8 & 36.0 & 61.6 \\
% CoG~\citep{chainofground} & - & - & - & - & - & - & - & - & - & - & - & - & 68.4 \\
UI-Zoomer~\citep{uizoomer} & 85.7 & 42.1 & 75.1 & 44.8 & 76.1 & 40.6 & 84.0 & 42.7 & 86.5 & 69.8 & 83.2 & 48.3 & 67.8 \\
ZoomClick~\citep{zoomclick} & 88.3 & 55.9 & 82.8 & 38.5 & 80.7 & 42.2 & 90.2 & 40.9 & 93.8 & 77.4 & 84.1 & 51.7 & 72.1 \\
MVP-8B~\citep{mvp} & - & - & - & - & - & - & - & - & - & - & - & - & 65.3 \\
MVP-32B~\citep{mvp} & - & - & - & - & - & - & - & - & - & - & - & - & 74.0 \\
\midrule

\textit{\textcolor{gray}{Qwen3-VL-8B-Instruct~\citep{qwen3vl}}}\repro 
& \textcolor{gray}{87.7} & \textcolor{gray}{37.9} & \textcolor{gray}{78.8} & \textcolor{gray}{36.4} & \textcolor{gray}{72.6} & \textcolor{gray}{28.1} & \textcolor{gray}{86.8} & \textcolor{gray}{38.2} & \textcolor{gray}{89.3} & \textcolor{gray}{54.7} & \textcolor{gray}{79.4} & \textcolor{gray}{47.2} & \textcolor{gray}{65.8} \\
\ \ + Re-Prefill         & 90.9 & 47.6 & 81.8 & 39.2 & 73.1 & 40.6 & 89.6 & 41.8 & 92.7 & 66.0 & 86.0 & 50.6 & 70.1\up{4.3} \\
\midrule

\textit{\textcolor{gray}{MAI-UI-8B~\citep{maiui}}}\repro 
& \textcolor{gray}{83.8} & \textcolor{gray}{49.0} & \textcolor{gray}{77.3} & \textcolor{gray}{40.6} & \textcolor{gray}{79.2} & \textcolor{gray}{34.4} & \textcolor{gray}{85.4} & \textcolor{gray}{48.2} & \textcolor{gray}{91.5} & \textcolor{gray}{69.8} & \textcolor{gray}{80.4} & \textcolor{gray}{56.2} & \textcolor{gray}{69.6} \\
\ \ + Re-Prefill         & 87.7 & 56.6 & 80.3 & 44.8 & 78.7 & 45.3 & 81.9 & 50.9 & 91.5 & 77.4 & 83.2 & 55.1 & 72.0\up{2.4} \\
\midrule

\textit{\textcolor{gray}{GUI-Owl-1.5-8B-Instruct~\citep{guiowl}}}\repro 
& \textcolor{gray}{81.2} & \textcolor{gray}{69.7} & \textcolor{gray}{83.8} & \textcolor{gray}{58.0} & \textcolor{gray}{84.8} & \textcolor{gray}{57.8} & \textcolor{gray}{93.1} & \textcolor{gray}{65.5} & \textcolor{gray}{93.2} & \textcolor{gray}{69.8} & \textcolor{gray}{85.1} & \textcolor{gray}{61.8} & \textcolor{gray}{78.0} \\
\ \ + Re-Prefill         & 88.3 & 73.8 & 85.9 & 58.0 & 85.8 & 64.1 & 91.0 & 65.5 & 93.2 & 75.5 & 86.9 & 67.4 & 80.1\up{2.1} \\
\midrule

\textit{\textcolor{gray}{Qwen3-VL-32B-Instruct~\citep{qwen3vl}}}\repro 
& \textcolor{gray}{89.6} & \textcolor{gray}{49.7} & \textcolor{gray}{85.4} & \textcolor{gray}{46.2} & \textcolor{gray}{83.8} & \textcolor{gray}{45.3} & \textcolor{gray}{89.6} & \textcolor{gray}{47.3} & \textcolor{gray}{94.4} & \textcolor{gray}{77.4} & \textcolor{gray}{91.6} & \textcolor{gray}{53.9} & \textcolor{gray}{74.3} \\
\ \ + Re-Prefill         & 93.5 & 54.5 & 87.4 & 51.7 & 83.8 & 43.8 & 94.4 & 49.1 & 94.9 & 77.4 & 88.8 & 64.0 & 76.8\up{2.5} \\
\bottomrule
\end{tabular}%
}
\vspace{-1.3em}
\end{table*}
\begin{table}[t]
\centering
\caption{Results on OSWorld-G. Methods marked with \repro{} are reproduced using the zoom-in strategy adopted in this work.}
\label{tab:main_osworld_classified}
\resizebox{\columnwidth}{!}{%
\begin{tabular}{l c c c c c >{\columncolor{blue!10}}c}
\toprule
\textbf{Model} & \textbf{Text Match.} & \textbf{Element Rec.} & \textbf{Layout Under.} & \textbf{Fine-grained Manip.} & \textbf{Refusal} & \multicolumn{1}{c}{\cellcolor{white}\textbf{Overall}} \\
\midrule
UI-TARS-72B~\citep{uitars} & 69.4 & 60.6 & 62.9 & 45.6 & - & 57.1 \\
UI-TARS-1.5-7B~\citep{uitars15} & 36.8 & 62.7 & 62.2 & 50.8 & - & 52.8 \\
% GTA1-7B~\citep{gta1} & 42.1 & 65.7 & 62.7 & 56.1 & - & 55.1 \\
GTA1-32B~\citep{gta1} & 63.2 & 78.4 & 73.3 & 65.2 & - & 65.2 \\
% GUI-Spotlight~\citep{guispotlight} & 68.2 & 60.6 & 63.2 & 45.6 & - & 62.7 \\
% OpenCUA-32B~\citep{opencua} & - & - & - & - & - & 59.6 \\
% EvoCUA-32B~\citep{evocua} & - & - & - & - & - & 63.9 \\
MAI-UI-8B~\citep{maiui} & 72.8 & 67.6 & 71.1 & 56.4 & - & 64.2 \\
GUI-Owl-1.5-8B-Instruct~\citep{guiowl} & 67.8 & 68.5 & 68.5 & 65.5 & 42.6 & 65.8 \\
\midrule

\textit{\textcolor{gray}{Qwen3-VL-8B-Instruct~\citep{qwen3vl}}}\repro & 
\textcolor{gray}{72.8} & \textcolor{gray}{66.7} & \textcolor{gray}{67.2} & \textcolor{gray}{53.7} & \textcolor{gray}{-} & \textcolor{gray}{63.0} \\
\ \ + Re-Prefill & 75.5 & 69.7 & 71.2 & 54.4 & - & 65.7\up{2.7} \\
\midrule

\textit{\textcolor{gray}{MAI-UI-8B~\citep{maiui}}}\repro & 
\textcolor{gray}{75.5} & \textcolor{gray}{70.3} & \textcolor{gray}{70.0} & \textcolor{gray}{53.7} & \textcolor{gray}{-} & \textcolor{gray}{65.5} \\
\ \ + Re-Prefill & 76.6 & 71.8 & 74.7 & 55.7 & - & 67.7\up{2.2} \\
\midrule

\textit{\textcolor{gray}{GUI-Owl-1.5-8B-Instruct~\citep{guiowl}}}\repro & 
\textcolor{gray}{75.5} & \textcolor{gray}{71.8} & \textcolor{gray}{73.5} & \textcolor{gray}{56.4} & \textcolor{gray}{-} & \textcolor{gray}{67.2} \\
\ \ + Re-Prefill & 77.4 & 73.0 & 74.3 & 59.7 & - & 68.8\up{1.6} \\
\midrule

\textit{\textcolor{gray}{Qwen3-VL-32B-Instruct~\citep{qwen3vl}}}\repro & 
\textcolor{gray}{76.6} & \textcolor{gray}{73.9} & \textcolor{gray}{77.9} & \textcolor{gray}{54.4} & \textcolor{gray}{-} & \textcolor{gray}{69.0} \\
\ \ + Re-Prefill & 78.9 & 74.8 & 77.5 & 57.0 & - & 70.1\up{1.1} \\
\bottomrule
\end{tabular}%
}
\vspace{-1.5em}
\end{table}

\begin{table}[t]
\centering
\caption{Results on ScreenSpot-V2. Methods marked with \repro{} are reproduced in this work.}
\label{tab:main_ssv2}
\footnotesize
\begin{tabular}{l *{6}{c} >{\columncolor{blue!10}}c}
\toprule
\multirow{2}{*}{\textbf{Model}} &
\multicolumn{2}{c}{\textbf{Mobile}} &
\multicolumn{2}{c}{\textbf{Desktop}} &
\multicolumn{2}{c}{\textbf{Web}} &
\multicolumn{1}{c}{\multirow{2}{*}{\cellcolor{white}\textbf{Overall}}} \\
\cmidrule(lr){2-3} \cmidrule(lr){4-5} \cmidrule(lr){6-7}
 & Text & Icon & Text & Icon & Text & Icon & \cellcolor{white} \\
\midrule
UI-TARS-72B~\citep{uitars} & 94.8 & 86.3 & 91.2 & 87.9 & 91.5 & 87.7 & 90.3 \\
% GTA1-7B~\citep{gta1} & 99.0 & 88.6 & 94.9 & 89.3 & 92.3 & 86.7 & 92.4 \\
GTA1-32B~\citep{gta1} & 99.7 & 90.5 & 99.0 & 94.3 & 95.7 & 90.1 & 95.2 \\
% OpenCUA-32B~\citep{opencua} & - & - & - & - & - & - & 93.4 \\
GUI-Cursor-7B~\citep{guicursor} & 99.2 & 90.6 & 94.4 & 91.3 & 96.1 & 89.0 & 93.9 \\
UI-Ins-32B~\citep{uiins} & 98.6 & 90.0 & 99.0 & 87.9 & 97.0 & 93.1 & 94.9 \\
GuirlVG~\citep{guirlvg} & 98.3 & 89.6 & 94.3 & 80.7 & 95.7 & 86.2 & 91.9 \\
% EvoCUA-32B~\citep{evocua} & - & - & - & - & - & - & 90.4 \\
GUI-Owl-1.5-8B-Instruct~\citep{guiowl} & 97.4 & 90.5 & 96.4 & 90.7 & 94.2 & 89.7 & 93.7 \\
\midrule

\textit{\textcolor{gray}{Qwen3-VL-8B-Instruct~\citep{qwen3vl}}}\repro &
\textcolor{gray}{99.3} & \textcolor{gray}{88.2} & \textcolor{gray}{97.9} & \textcolor{gray}{87.9} & \textcolor{gray}{96.6} & \textcolor{gray}{87.2} & \textcolor{gray}{93.6} \\
\ \ + Re-Prefill & 99.7 & 89.6 & 97.9 & 90.0 & 97.0 & 90.1 & 94.7\up{1.1} \\
\midrule

\textit{\textcolor{gray}{MAI-UI-8B~\citep{maiui}}}\repro &
\textcolor{gray}{99.0} & \textcolor{gray}{89.6} & \textcolor{gray}{97.9} & \textcolor{gray}{91.4} & \textcolor{gray}{97.4} & \textcolor{gray}{91.6} & \textcolor{gray}{94.2} \\
\ \ + Re-Prefill & 99.3 & 90.0 & 97.9 & 92.1 & 97.9 & 91.6 & 95.3\up{1.1} \\
\midrule

\textit{\textcolor{gray}{GUI-Owl-1.5-8B-Instruct~\citep{guiowl}}}\repro &
\textcolor{gray}{98.6} & \textcolor{gray}{90.0} & \textcolor{gray}{97.4} & \textcolor{gray}{89.3} & \textcolor{gray}{96.2} & \textcolor{gray}{88.2} & \textcolor{gray}{93.9} \\
\ \ + Re-Prefill & 98.3 & 91.0 & 97.9 & 92.1 & 96.2 & 89.7 & 94.6\up{0.7} \\
\midrule

\textit{\textcolor{gray}{Qwen3-VL-32B-Instruct~\citep{qwen3vl}}}\repro &
\textcolor{gray}{99.0} & \textcolor{gray}{91.5} & \textcolor{gray}{98.5} & \textcolor{gray}{90.0} & \textcolor{gray}{96.6} & \textcolor{gray}{91.1} & \textcolor{gray}{95.0} \\
\ \ + Re-Prefill & 99.3 & 91.9 & 98.5 & 91.4 & 97.4 & 90.1 & 95.3\up{0.3} \\
\bottomrule
\end{tabular}
\vspace{-2.6em}
\end{table}

\noindent{\textbf{High-resolution and dense UI grounding. }}
ScreenSpot-Pro evaluates grounding on high-resolution interfaces with dense layouts.
As shown in Table~\ref{tab:main_sspro}, Re-Prefill consistently improves all four models, with gains of $+4.3$ (Qwen3-VL-8B), $+2.4$ (MAI-UI-8B), $+2.1$ (GUI-Owl-8B), and $+2.5$ (Qwen3-VL-32B).
For Qwen3-VL models, performance increases from $65.8\%$ to $70.1\%$ (8B) and from $74.3\%$ to $76.8\%$ (32B), outperforming MVP ($65.3\%$ and $74.0\%$) under the same backbone.
For GUI-specialized models, Re-Prefill also yields consistent gains, improving MAI-UI-8B from $69.6\%$ to $72.0\%$ and GUI-Owl-1.5-8B from $78.0\%$ to $80.1\%$, despite prior GUI-specific training.
These results show that Re-Prefill complements training-based approaches by refining target selection at inference time.
The larger gains on general-purpose models suggest that improving the prefill stage is particularly beneficial when initial target selection is less reliable.

\noindent{\textbf{Complex desktop interaction grounding. }}
OSWorld-G evaluates grounding in desktop environments requiring structural UI understanding and interaction-aware reasoning. It provides original and refined annotation versions. This section reports results on the original annotations, while the refined results are provided in Appendix~\ref{app:osworldg_refined}.
As shown in Table~\ref{tab:main_osworld_classified}, Re-Prefill improves all models, with gains of $+2.7$ (Qwen3-VL-8B), $+2.2$ (MAI-UI-8B), $+1.6$ (GUI-Owl-8B), and $+1.1$ (Qwen3-VL-32B).
Compared with ScreenSpot-Pro, improvements are smaller but more stable across model scales.
Qwen3-VL improves from $63.0\%$ to $65.7\%$ (8B) and from $69.0\%$ to $70.1\%$ (32B), while MAI-UI-8B and GUI-Owl-1.5-8B increase from $65.5\%$ to $67.7\%$ and from $67.2\%$ to $68.8\%$, respectively.
These results indicate that Re-Prefill remains effective under complex interaction settings, improving both visual discrimination and structural reasoning.
The smaller gains suggest that when grounding relies more on interaction structure than visual ambiguity, the benefit of prefill refinement becomes less pronounced but remains consistent.

\noindent{\textbf{Cross-platform UI grounding. }}
ScreenSpot-V2 evaluates grounding across mobile, desktop, and web interfaces.
As shown in Table~\ref{tab:main_ssv2}, Re-Prefill improves all models, with gains of $+1.1$ (Qwen3-VL-8B), $+1.1$ (MAI-UI-8B), $+0.7$ (GUI-Owl-8B), and $+0.3$ (Qwen3-VL-32B).
Improvements are smaller than on ScreenSpot-Pro and OSWorld-G due to the already high baseline performance (above $93\%$), leaving limited room for improvement.
Despite this saturation, gains remain consistent across model types and platforms, including both mobile and desktop interfaces.
These results suggest that Re-Prefill generalizes well across diverse UI domains, acting as a robust inference-time enhancement even when performance is near saturation.

\begin{wraptable}{r}{0.54\textwidth}
\vspace{-1em}
\centering
\caption{Results on UI-Vision. Methods marked with \repro{} are reproduced using the zoom-in strategy adopted in this work.}
\label{tab:main_uivision}
\scriptsize
\setlength{\tabcolsep}{3pt}

\begin{tabularx}{\linewidth}{
l 
>{\centering\arraybackslash}X 
>{\centering\arraybackslash}X 
>{\centering\arraybackslash}X 
>{\columncolor{blue!10}\centering\arraybackslash}X
}
\toprule
\textbf{Model} & \textbf{Basic} & \textbf{Functional} & \textbf{Spatial} & \textbf{Overall} \\
\midrule
UI-TARS-72B~\citep{uitars} & 31.4 & 30.5 & 14.7 & 25.5 \\
UI-TARS-1.5-7B~\citep{uitars15} & 28.8 & 27.5 & 10.7 & 22.3 \\
MAI-UI-8B~\citep{maiui} & 51.6 & 50.5 & 26.6 & 42.4 \\
\midrule
\multicolumn{5}{l}{\textit{Training-free methods}} \\
UI-Zoomer~\citep{uizoomer} & 42.5 & 40.1 & 19.7 & 33.7 \\
ZoomClick~\citep{zoomclick} & 45.1 & 45.1 & 28.3 & 39.2 \\
MVP-8B~\citep{mvp} & 38.3 & 38.8 & 18.7 & 31.9 \\
MVP-32B~\citep{mvp} & 49.4 & 52.0 & 30.8 & 44.1 \\
\midrule

\textit{\textcolor{gray}{Qwen3-VL-8B-Instruct~\citep{qwen3vl}}}\repro & 
\textcolor{gray}{38.2} & \textcolor{gray}{39.9} & \textcolor{gray}{19.3} & \textcolor{gray}{32.1} \\
\ \ + Re-Prefill & 39.4 & 41.7 & 19.3 & 33.1\up{1.0} \\
\midrule

\textit{\textcolor{gray}{MAI-UI-8B~\citep{maiui}}}\repro & 
\textcolor{gray}{50.7} & \textcolor{gray}{50.4} & \textcolor{gray}{27.0} & \textcolor{gray}{42.2} \\
\ \ + Re-Prefill & 52.3 & 52.8 & 27.0 & 43.5\up{1.3} \\
\midrule

\textit{\textcolor{gray}{GUI-Owl-1.5-8B-Instruct~\citep{guiowl}}}\repro & 
\textcolor{gray}{50.1} & \textcolor{gray}{44.7} & \textcolor{gray}{29.8} & \textcolor{gray}{41.2} \\
\ \ + Re-Prefill & 50.1 & 46.1 & 29.7 & 41.6\up{0.4} \\
\midrule

\textit{\textcolor{gray}{Qwen3-VL-32B-Instruct~\citep{qwen3vl}}}\repro & 
\textcolor{gray}{49.9} & \textcolor{gray}{51.1} & \textcolor{gray}{31.1} & \textcolor{gray}{43.7} \\
\ \ + Re-Prefill & 51.4 & 51.0 & 32.4 & 44.6\up{0.9} \\
\bottomrule
\end{tabularx}
\vspace{-1em}
\end{wraptable}

\noindent{\textbf{Instruction-diverse grounding. }}
We further evaluate Re-Prefill on instruction-heavy grounding scenarios using UI-Vision and MMBench-GUI-L2, which require robust query understanding under varying levels of semantic abstraction.
On UI-Vision (Table~\ref{tab:main_uivision}), Re-Prefill improves all models, with gains of $+1.0$ (Qwen3-VL-8B), $+1.3$ (MAI-UI-8B), $+0.4$ (GUI-Owl-8B), and $+0.9$ (Qwen3-VL-32B). 
On MMBench-GUI-L2 (Table~\ref{tab:main_mmbench}), the gains increase to $+2.2$, $+1.3$, $+0.6$, and $+1.1$, respectively.
The consistent improvements across both benchmarks indicate that Re-Prefill remains effective under diverse instructions.
Compared with scenarios dominated by visual ambiguity or structural complexity, these results suggest that refining the prefill stage also benefits semantic alignment between instructions and UI elements, enabling more robust grounding under varying query formulations.

\begin{table*}[t]
\centering
\caption{Results on MMBench-GUI-L2. Methods marked with \repro{} are reproduced using the zoom-in strategy adopted in this work.}
\label{tab:main_mmbench}
\resizebox{\textwidth}{!}{%
\begin{tabular}{l *{12}{c} >{\columncolor{blue!10}}c}
\toprule
\multirow{2}{*}{\textbf{Model}} &
\multicolumn{2}{c}{\textbf{Windows}} &
\multicolumn{2}{c}{\textbf{MacOS}} &
\multicolumn{2}{c}{\textbf{Linux}} &
\multicolumn{2}{c}{\textbf{iOS}} &
\multicolumn{2}{c}{\textbf{Android}} &
\multicolumn{2}{c}{\textbf{Web}} &
\multicolumn{1}{c}{\multirow{2}{*}{\cellcolor{white}\textbf{Overall}}} \\
\cmidrule(lr){2-3} \cmidrule(lr){4-5} \cmidrule(lr){6-7} \cmidrule(lr){8-9} \cmidrule(lr){10-11} \cmidrule(lr){12-13}
 & Basic & Adv. & Basic & Adv. & Basic & Adv. & Basic & Adv. & Basic & Adv. & Basic & Adv. & \cellcolor{white} \\
\midrule
UI-TARS-72B~\citep{uitars} & 78.6 & 51.8 & 80.3 & 62.7 & 68.6 & 51.5 & 90.8 & 81.2 & 93.0 & 80.0 & 88.1 & 68.5 & 74.3 \\
UI-TARS-1.5-7B~\citep{uitars15} & 68.3 & 39.0 & 69.0 & 44.5 & 64.4 & 37.8 & 88.5 & 69.4 & 90.5 & 69.3 & 81.0 & 56.5 & 64.3 \\
UI-Ins-32B~\citep{uiins} & 84.9 & 68.4 & 88.4 & 73.4 & 68.6 & 56.1 & 96.5 & 91.2 & 97.2 & 92.4 & 94.8 & 85.1 & 84.9 \\
MAI-UI-8B~\citep{maiui} & 92.3 & 74.3 & 90.7 & 86.4 & 81.2 & 67.3 & 97.1 & 90.0 & 97.5 & 92.7 & 95.8 & 86.0 & 88.8 \\
GUI-Owl-1.5-8B-Instruct~\citep{guiowl} & 89.7 & 65.4 & 88.1 & 72.8 & 72.8 & 56.6 & 95.9 & 83.9 & 95.2 & 82.9 & 93.2 & 77.6 & 82.5 \\
\midrule

\textit{\textcolor{gray}{Qwen3-VL-8B-Instruct~\citep{qwen3vl}}}\repro &
\textcolor{gray}{90.4} & \textcolor{gray}{70.6} & \textcolor{gray}{86.7} & \textcolor{gray}{73.4} & \textcolor{gray}{77.5} & \textcolor{gray}{62.8} & \textcolor{gray}{95.2} & \textcolor{gray}{85.8} & \textcolor{gray}{95.5} & \textcolor{gray}{86.8} & \textcolor{gray}{94.8} & \textcolor{gray}{78.2} & \textcolor{gray}{84.2} \\
\ \ + Re-Prefill & 90.0 & 73.5 & 89.0 & 78.3 & 82.2 & 67.9 & 96.2 & 86.7 & 96.1 & 88.5 & 95.5 & 82.1 & 86.4\up{2.2} \\
\midrule

\textit{\textcolor{gray}{MAI-UI-8B~\citep{maiui}}}\repro &
\textcolor{gray}{93.0} & \textcolor{gray}{71.0} & \textcolor{gray}{91.9} & \textcolor{gray}{78.6} & \textcolor{gray}{83.2} & \textcolor{gray}{67.9} & \textcolor{gray}{96.5} & \textcolor{gray}{87.6} & \textcolor{gray}{98.0} & \textcolor{gray}{88.2} & \textcolor{gray}{96.4} & \textcolor{gray}{83.1} & \textcolor{gray}{87.2} \\
\ \ + Re-Prefill & 93.0 & 74.6 & 93.0 & 80.6 & 83.8 & 71.4 & 95.9 & 90.0 & 97.8 & 91.3 & 95.5 & 84.4 & 88.5\up{1.3} \\
\midrule

\textit{\textcolor{gray}{GUI-Owl-1.5-8B-Instruct~\citep{guiowl}}}\repro &
\textcolor{gray}{92.6} & \textcolor{gray}{65.8} & \textcolor{gray}{88.4} & \textcolor{gray}{73.1} & \textcolor{gray}{82.2} & \textcolor{gray}{58.2} & \textcolor{gray}{97.1} & \textcolor{gray}{86.4} & \textcolor{gray}{97.5} & \textcolor{gray}{83.9} & \textcolor{gray}{95.8} & \textcolor{gray}{80.2} & \textcolor{gray}{84.5} \\
\ \ + Re-Prefill & 90.8 & 68.8 & 87.8 & 74.3 & 82.7 & 61.7 & 96.5 & 86.7 & 97.8 & 85.1 & 95.5 & 81.2 & 85.1\up{0.6} \\
\midrule

\textit{\textcolor{gray}{Qwen3-VL-32B-Instruct~\citep{qwen3vl}}}\repro &
\textcolor{gray}{93.4} & \textcolor{gray}{71.3} & \textcolor{gray}{90.4} & \textcolor{gray}{80.3} & \textcolor{gray}{86.9} & \textcolor{gray}{70.4} & \textcolor{gray}{97.1} & \textcolor{gray}{90.3} & \textcolor{gray}{97.8} & \textcolor{gray}{91.5} & \textcolor{gray}{96.8} & \textcolor{gray}{84.1} & \textcolor{gray}{88.4} \\
\ \ + Re-Prefill & 92.6 & 72.4 & 91.9 & 84.7 & 85.9 & 76.0 & 97.5 & 91.8 & 96.9 & 91.5 & 97.1 & 86.4 & 89.5\up{1.1} \\
\bottomrule
\end{tabular}%
}
\vspace{-0.8em}
\end{table*}

\begin{figure}[!t]
  % \vspace{-0.8em}
  \centering
  \includegraphics[width=\textwidth]{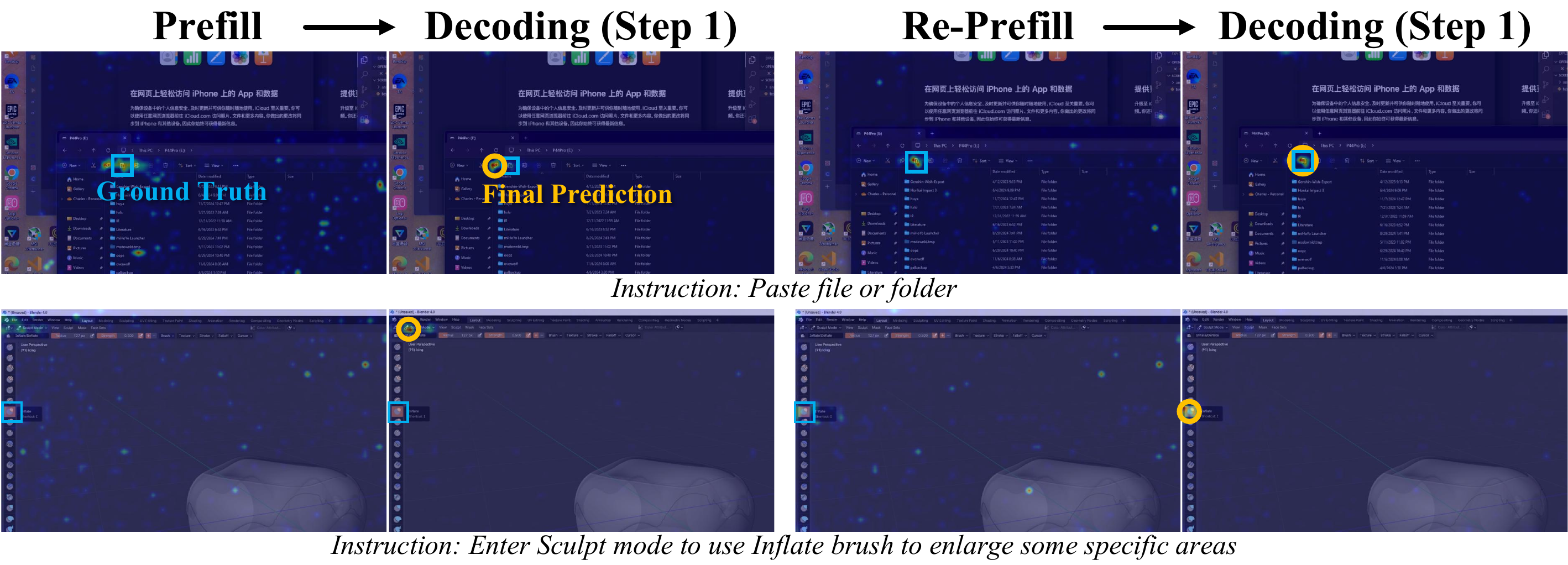}
  \caption{
    \textbf{Query-position attention heatmaps across stages on ScreenSpot-Pro.}
    The first two panels illustrate the baseline transition, while the last two panels show the corresponding transition under Re-Prefill.
    The blue rectangle marks the ground-truth target, and the orange circle indicates the predicted coordinate.
    Re-Prefill focuses attention on the correct region during prefill, suppresses distractors, and leads to more accurate localization in subsequent decoding.
  }
  \label{fig:attn_visualization_1}
  \vspace{-1.8em}
\end{figure}

\noindent{\textbf{Qualitative analysis. }}
Figure~\ref{fig:attn_visualization_1} visualizes query-position attention across stages for a representative 

\noindent example, comparing the baseline with Re-Prefill. In the baseline, attention at prefill is dispersed over multiple candidate regions and may contract onto an incorrect region during decoding.
In contrast, Re-Prefill produces more focused attention on the correct target during prefill, leading to more accurate localization in subsequent decoding.
This observation aligns with the two-stage inference structure in Section~\ref{sec:bottleneck}, where target selection is determined during prefill and propagated through decoding.
Additional qualitative results are provided in Appendix~\ref{app:attn_visualization}.

Overall, Re-Prefill consistently improves grounding accuracy across diverse settings.
A detailed analysis of computational efficiency is provided in Appendix~\ref{app:efficiency}.

\subsection{Ablation Studies}
\label{sec:ablation}

\begin{wraptable}{r}{0.48\linewidth}
    \vspace{-3.5em}
    \centering
    \caption{Analysis of Re-Prefill mechanism. Embedding Addition removes the second prefill and uses the first-prefill visual tokens in decoding. Blind Re-Prefill omits selection and uses all visual tokens. Random Token Selection replaces attention-based selection with random sampling.}
    \footnotesize
    \begin{tabular}{lc}
    \toprule
    \textbf{Method} & \textbf{Acc.} \\
    \midrule
    Baseline & 65.8 \\
    \midrule
    Embedding Addition       & 66.0 \\
    Blind Re-Prefill         & 69.4 \\
    Random Token Selection   & 69.1 \\
    \textbf{Re-Prefill (Ours)} & \textbf{70.1} \\
    \bottomrule
    \end{tabular}
    \label{tab:ablation_component}
    \vspace{-2.0em}
\end{wraptable}

Ablation studies are conducted on ScreenSpot-Pro with Qwen3-VL-8B-Instruct as the base model.

% -------- 4.3.1 --------
\noindent{\textbf{Analysis of the Re-Prefill mechanism. }}
Three variants are designed to verify the contribution of the Re-Prefill mechanism (Table~\ref{tab:ablation_component}).
\emph{Embedding Addition} reuses the first-prefill visual tokens $\tilde{\mathbf{V}}$ in the decoding context instead of the enriched $\hat{\mathbf{V}}$, thereby skipping the second prefill stage.
\emph{Blind Re-Prefill} omits key-token selection and appends the full contextualized visual tokens $\tilde{\mathbf{V}}$ in the second prefill.
\emph{Random Token Selection} replaces attention-based token selection with random sampling.
All three variants perform worse than Re-Prefill.
The largest gap, observed for Embedding Addition ($66.0\%$), confirms that the primary source of improvement is the second prefill stage. The benefit arises from allowing the original visual tokens to undergo a second prefill guided by contextualized key visual tokens and instruction tokens, rather than merely emphasizing visual information from the first pass.
The gap to Blind Re-Prefill ($69.4\%$) indicates that focusing the second pass on a compact set of key visual tokens is more effective than propagating the full sequence.
The gap to Random Token Selection ($69.1\%$) shows that the attention-derived selection signal further enhance performance by filtering out unrelated tokens and reducing interference from noisy information.

\begin{figure}[t]
    \vspace{-1em}
    \centering
    %------------- RIGHT: Figure 4 (rho-gamma heatmap) -------------
    \begin{minipage}[t]{0.48\linewidth}
        \centering
        \includegraphics[width=\linewidth]{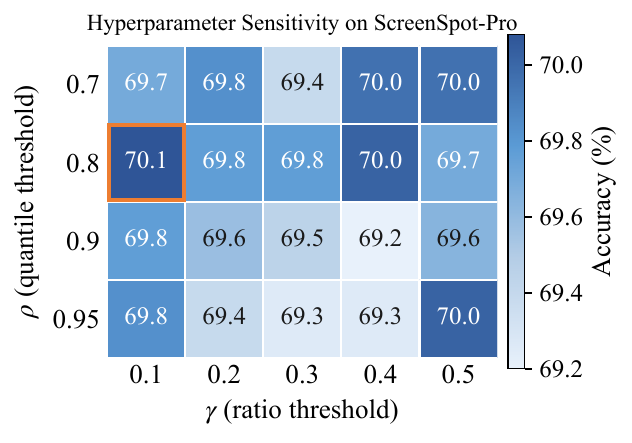}
        \captionof{figure}{Effect of $\rho$ and $\gamma$. The orange box marks the default configuration ($\rho{=}0.8$, $\gamma{=}0.1$). All twenty cells exceed the baseline ($65.8\%$) with a spread of less than one point, indicating that Re-Prefill is largely insensitive to these thresholds.}
        \label{fig:ablation_hyper}
        \vspace{-2em}
    \end{minipage}
    \hfill
     %------------- LEFT: Figure 5 (L_c ablation) -------------
    \begin{minipage}[t]{0.48\linewidth}
        \centering
        \includegraphics[width=\linewidth]{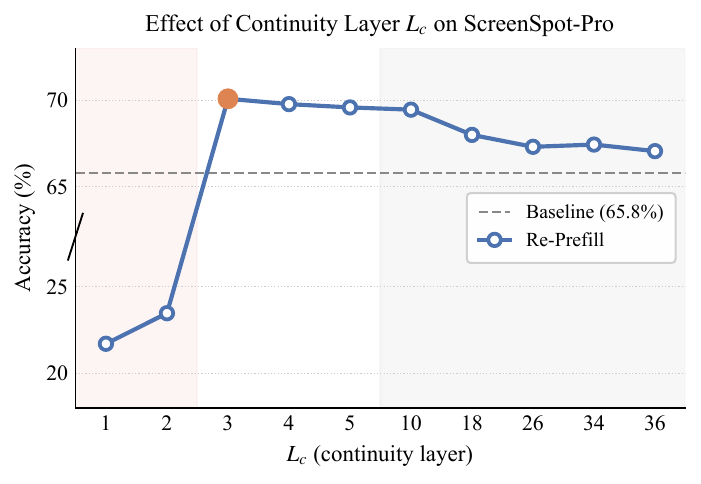}
        \captionof{figure}{Effect of $L_c$. The optimum at $L_c{=}3$ balances two modes. For small $L_c$, insufficient semantic alignment arises between uncontextualized input tokens and the first-prefill prefix (red zone). For large $L_c$, noise from unrelated tokens propagates into deeper layers (grey zone).}
        \label{fig:ablation_lc}
        \vspace{-0.3em}
    \end{minipage}
    \vspace{-1.0em}
\end{figure}

% -------- 4.3.2 --------
\noindent{\textbf{Analysis of hyperparameters $\rho$ and $\gamma$. }}
The selection of $\tilde{\mathbf{V}}^{*}$ depends on the quantile threshold $\rho$ (Eq.~\ref{eq:threshold}) and the ratio threshold $\gamma$ (Eq.~\ref{eq:ratio}).
Figure~\ref{fig:ablation_hyper} presents a $4 \times 5$ grid over $\rho \in \{0.70, 0.80, 0.90, 0.95\}$ and $\gamma \in \{0.1, 0.2, 0.3, 0.4, 0.5\}$.
All twenty configurations outperform the $65.8\%$ baseline, and the variation across the grid remains within $1.0$ point ($69.2\%$ to $70.1\%$), indicating that Re-Prefill is largely insensitive to these thresholds.
The best configuration ($\rho = 0.8$, $\gamma = 0.1$) lies at the more permissive end of both axes, expanding $\tilde{\mathbf{V}}^{*}$ to include not only the most highly attended region but also neighboring candidates that compete for attention.
This design allows the second prefill stage to recover when the initial selection corresponds to an incorrect UI element.
This configuration is adopted in all main results.

% -------- 4.3.3 --------
\noindent{\textbf{Analysis of continuity layers $L_c$. }}
$L_c$ determines how many early decoder layers retain the full prefix from the first prefill before it is reduced to the key visual tokens $\tilde{\mathbf{V}}^{*}$ and instruction tokens $\tilde{\mathbf{T}}$.
A sweep of $L_c$ from $1$ to $36$ is conducted on Qwen3-VL-8B-Instruct ($L = 36$).
Figure~\ref{fig:ablation_lc} shows that the accuracy drops to approximately $22\%$ for $L_c \in \{1,2\}$, reaches a maximum of $70.1\%$ at $L_c = 3$, and then decreases monotonically.
This behavior reflects the dual role of $L_c$. Early layers ($\le L_c$) use the full contextualized representations from the first prefill as a prefix to align the copy of uncontextualized input tokens with it. Later layers ($> L_c$) retain only $\tilde{\mathbf{V}}^{*}$ and $\tilde{\mathbf{T}}$ so that subsequent processing concentrates on candidate regions without interference from unrelated tokens.
When $L_c$ is small, the prefix is truncated before semantic alignment is established, causing later layers to operate on uncalibrated inputs with incomplete cues, which leads to severe performance degradation.
When $L_c$ is large, unrelated tokens propagate into deeper layers and dilute attention away from candidate regions, thereby reducing accuracy.
The optimal value at $L_c = 3$ balances these two failure modes.

\section{Conclusion}
This work analyzes the inference dynamics of decoder-only VLMs for GUI grounding and identifies a two-stage paradigm in which target selection is completed during the prefill stage, while decoding only refines coordinates within the selected region.
The prefill stage therefore constitutes the decisive factor for grounding accuracy, yet it remains unaddressed by existing multi-run methods.
Motivated by this observation, Re-Prefill is proposed as a training-free approach that calibrates the prefill stage through attention-guided re-evaluation, in which key visual tokens and instruction tokens are reintroduced for a second prefill stage prior to coordinate decoding.
Across four VLMs and five benchmarks, consistent improvements are achieved without additional training, reaching gains of up to 4.3 points on ScreenSpot-Pro and 2.7 points on OSWorld-G.
These results establish prefill calibration as a principled direction for inference-time computation in vision-language models.

\bibliographystyle{unsrtnat}
\bibliography{references}

@article{qwen3vl,
  title={Qwen3-vl technical report},
  author={Bai, Shuai and Cai, Yuxuan and Chen, Ruizhe and Chen, Keqin and Chen, Xionghui and Cheng, Zesen and Deng, Lianghao and Ding, Wei and Gao, Chang and Ge, Chunjiang and others},
  journal={arXiv preprint arXiv:2511.21631},
  year={2025}
}

@inproceedings{survey_nguyen2025gui,
  title={Gui agents: A survey},
  author={Nguyen, Dang and Chen, Jian and Wang, Yu and Wu, Gang and Park, Namyong and Hu, Zhengmian and Lyu, Hanjia and Wu, Junda and Aponte, Ryan and Xia, Yu and others},
  booktitle={Findings of the Association for Computational Linguistics: ACL 2025},
  pages={22522--22538},
  year={2025}
}

@article{survey_zhang2024large,
  title={Large language model-brained gui agents: A survey},
  author={Zhang, Chaoyun and He, Shilin and Qian, Jiaxu and Li, Bowen and Li, Liqun and Qin, Si and Kang, Yu and Ma, Minghua and Liu, Guyue and Lin, Qingwei and others},
  journal={arXiv preprint arXiv:2411.18279},
  year={2024}
}

@article{survey_wang2024gui,
  title={Gui agents with foundation models: A comprehensive survey},
  author={Wang, Shuai and Liu, Weiwen and Chen, Jingxuan and Zhou, Yuqi and Gan, Weinan and Zeng, Xingshan and Che, Yuhan and Yu, Shuai and Hao, Xinlong and Shao, Kun and others},
  journal={arXiv preprint arXiv:2411.04890},
  year={2024}
}

@article{survey_tang2025survey,
  title={A survey on (m) llm-based gui agents},
  author={Tang, Fei and Xu, Haolei and Zhang, Hang and Chen, Siqi and Wu, Xingyu and Shen, Yongliang and Zhang, Wenqi and Hou, Guiyang and Tan, Zeqi and Yan, Yuchen and others},
  journal={arXiv preprint arXiv:2504.13865},
  year={2025}
}

@inproceedings{guig2,
  title={GUI-G$^2$: Gaussian Reward Modeling for GUI Grounding},
  author={Tang, Fei and Gu, Zhangxuan and Lu, Zhengxi and Liu, Xuyang and Shen, Shuheng and Meng, Changhua and Wang, Wen and Zhang, Wenqi and Shen, Yongliang and Lu, Weiming and others},
  booktitle={Proceedings of the AAAI Conference on Artificial Intelligence},
  volume={40},
  number={39},
  pages={33214--33222},
  year={2026}
}

@article{uitars,
  title={Ui-tars: Pioneering automated gui interaction with native agents},
  author={Qin, Yujia and Ye, Yining and Fang, Junjie and Wang, Haoming and Liang, Shihao and Tian, Shizuo and Zhang, Junda and Li, Jiahao and Li, Yunxin and Huang, Shijue and others},
  journal={arXiv preprint arXiv:2501.12326},
  year={2025}
}

@misc{uitars15,
  title        = {UI-TARS-1.5: A Multimodal UI Understanding and Reasoning Model},
  author       = {{ByteDance Seed Team}},
  year         = {2025},
  organization = {ByteDance},
  url          = {https://seed-tars.com/1.5}
}

@article{guiowl,
  title={Mobile-Agent-v3. 5: Multi-platform Fundamental GUI Agents},
  author={Xu, Haiyang and Zhang, Xi and Liu, Haowei and Wang, Junyang and Zhu, Zhaozai and Zhou, Shengjie and Hu, Xuhao and Gao, Feiyu and Cao, Junjie and Wang, Zihua and others},
  journal={arXiv preprint arXiv:2602.16855},
  year={2026}
}

@article{maiui,
  title={MAI-UI Technical Report: Real-World Centric Foundation GUI Agents},
  author={Zhou, Hanzhang and Zhang, Xu and Tong, Panrong and Zhang, Jianan and Chen, Liangyu and Kong, Quyu and Cai, Chenglin and Liu, Chen and Wang, Yue and Zhou, Jingren and others},
  journal={arXiv preprint arXiv:2512.22047},
  year={2025}
}

@inproceedings{opencua,
  title={OpenCUA: Open Foundations for Computer-Use Agents},
  author={Wang, Xinyuan and Wang, Bowen and Lu, Dunjie and Yang, Junlin and Xie, Tianbao and Wang, Junli and Deng, Jiaqi and Guo, Xiaole and Xu, Yiheng and Wu, Chen Henry and others},
  booktitle={The Thirty-ninth Annual Conference on Neural Information Processing Systems},
  year={2025}
}

@article{evocua,
  title={Evocua: Evolving computer use agents via learning from scalable synthetic experience},
  author={Xue, Taofeng and Peng, Chong and Huang, Mianqiu and Guo, Linsen and Han, Tiancheng and Wang, Haozhe and Wang, Jianing and Zhang, Xiaocheng and Yang, Xin and Zhao, Dengchang and others},
  journal={arXiv preprint arXiv:2601.15876},
  year={2026}
}

@inproceedings{seeclick,
  title={Seeclick: Harnessing gui grounding for advanced visual gui agents},
  author={Cheng, Kanzhi and Sun, Qiushi and Chu, Yougang and Xu, Fangzhi and YanTao, Li and Zhang, Jianbing and Wu, Zhiyong},
  booktitle={Proceedings of the 62nd Annual Meeting of the Association for Computational Linguistics (Volume 1: Long Papers)},
  pages={9313--9332},
  year={2024}
}

@inproceedings{cogagent,
  title={Cogagent: A visual language model for gui agents},
  author={Hong, Wenyi and Wang, Weihan and Lv, Qingsong and Xu, Jiazheng and Yu, Wenmeng and Ji, Junhui and Wang, Yan and Wang, Zihan and Dong, Yuxiao and Ding, Ming and others},
  booktitle={Proceedings of the IEEE/CVF conference on computer vision and pattern recognition},
  pages={14281--14290},
  year={2024}
}

@inproceedings{ariaui,
  title={Aria-ui: Visual grounding for gui instructions},
  author={Yang, Yuhao and Wang, Yue and Li, Dongxu and Luo, Ziyang and Chen, Bei and Huang, Chao and Li, Junnan},
  booktitle={Findings of the Association for Computational Linguistics: ACL 2025},
  pages={22418--22433},
  year={2025}
}

@inproceedings{uground,
  title={Navigating the Digital World as Humans Do: Universal Visual Grounding for {GUI} Agents},
  author={Boyu Gou and Ruohan Wang and Boyuan Zheng and Yanan Xie and Cheng Chang and Yiheng Shu and Huan Sun and Yu Su},
  booktitle={The Thirteenth International Conference on Learning Representations},
  year={2025},
  url={https://openreview.net/forum?id=kxnoqaisCT}
}

@inproceedings{osatlas, 
  title={{OS}-{ATLAS}: Foundation Action Model for Generalist {GUI} Agents},
  author={Zhiyong Wu and Zhenyu Wu and Fangzhi Xu and Yian Wang and Qiushi Sun and Chengyou Jia and Kanzhi Cheng and Zichen Ding and Liheng Chen and Paul Pu Liang and Yu Qiao},
  booktitle={The Thirteenth International Conference on Learning Representations},
  year={2025},
  url={https://openreview.net/forum?id=n9PDaFNi8t}
}

@inproceedings{showui,
  title={Showui: One vision-language-action model for generalist gui agent},
  author={Lin, Kevin Qinghong and Li, Linjie and Gao, Difei and Yang, Zhengyuan and Bai, Zechen and Lei, Weixian and Wang, Lijuan and Shou, Mike Zheng},
  booktitle={NeurIPS 2024 Workshop on Open-World Agents},
  year={2024}
}

@article{guicursor,
  title={Learning gui grounding with spatial reasoning from visual feedback},
  author={Zhao, Yu and Chen, Wei-Ning and Inan, Huseyin Atahan and Kessler, Samuel and Wang, Lu and Wutschitz, Lukas and Yang, Fangkai and Zhang, Chaoyun and Minervini, Pasquale and Rajmohan, Saravan and others},
  journal={arXiv preprint arXiv:2509.21552},
  year={2025}
}

@inproceedings{gta1,
    title={{GTA}1: {GUI} Test-time Scaling Agent},
    author={Yan Yang and Dongxu Li and Yutong Dai and Yuhao Yang and Ziyang Luo and Zirui Zhao and Zhiyuan Hu and Junzhe Huang and Amrita Saha and Zeyuan Chen and Ran Xu and Liyuan Pan and Caiming Xiong and Junnan Li},
    booktitle={The Fourteenth International Conference on Learning Representations},
    year={2026},
    url={https://openreview.net/forum?id=3VIPmz7iAi}
}

@inproceedings{uiins,
    title={{UI}-Ins: Enhancing {GUI} Grounding with Multi-Perspective Instruction as Reasoning},
    author={Liangyu Chen and Hanzhang Zhou and Chenglin Cai and Jianan Zhang and Panrong Tong and Xu Zhang and Quyu Kong and Chen Liu and Yuqi Liu and Wenxuan Wang and Yue Wang and Qin Jin and Steven HOI},
    booktitle={The Fourteenth International Conference on Learning Representations},
    year={2026},
    url={https://openreview.net/forum?id=dsQHm7YX9c}
}

@inproceedings{guirlvg,
    title={Guirl{VG}: Incentivize {GUI} Visual Grounding via Empirical Exploration on Reinforcement Learning},
    author={Weitai Kang and Bin Lei and Gaowen Liu and Caiwen Ding and Yan Yan},
    booktitle={The Fourteenth International Conference on Learning Representations},
    year={2026},
    url={https://openreview.net/forum?id=zrH2A1upAo}
}

@article{guir1,
  title={Gui-r1: A generalist r1-style vision-language action model for gui agents},
  author={Luo, Run and Wang, Lu and He, Wanwei and Chen, Longze and Li, Jiaming and Xia, Xiaobo},
  journal={arXiv preprint arXiv:2504.10458},
  year={2025}
}

@article{yuan2025enhancing,
  title={Enhancing visual grounding for gui agents via self-evolutionary reinforcement learning},
  author={Yuan, Xinbin and Zhang, Jian and Li, Kaixin and Cai, Zhuoxuan and Yao, Lujian and Chen, Jie and Wang, Enguang and Hou, Qibin and Chen, Jinwei and Jiang, Peng-Tao and others},
  journal={arXiv preprint arXiv:2505.12370},
  year={2025}
}

@article{uivenus,
  title={Ui-venus technical report: Building high-performance ui agents with rft},
  author={Gu, Zhangxuan and Zeng, Zhengwen and Xu, Zhenyu and Zhou, Xingran and Shen, Shuheng and Liu, Yunfei and Zhou, Beitong and Meng, Changhua and Xia, Tianyu and Chen, Weizhi and others},
  journal={arXiv preprint arXiv:2508.10833},
  year={2025}
}

@article{guieyes,
  title={GUI-Eyes: Tool-Augmented Perception for Visual Grounding in GUI Agents},
  author={Chen, Chen and Shao, Jiawei and Lu, Dakuan and Hu, Haoyi and Liu, Xiangcheng and Yao, Hantao and Liu, Wu},
  journal={arXiv preprint arXiv:2601.09770},
  year={2026}
}

@article{mvp,
  title={MVP: Multiple View Prediction Improves GUI Grounding},
  author={Zhang, Yunzhu and Pan, Zeyu and Zeng, Zhengwen and Shen, Shuheng and Meng, Changhua and Zhu, Linchao},
  journal={arXiv preprint arXiv:2512.08529},
  year={2025}
}

@article{zoomclick,
  title={Zoom in, Click out: Unlocking and Evaluating the Potential of Zooming for GUI Grounding},
  author={Jiang, Zhiyuan and Xie, Shenghao and Li, Wenyi and Zu, Wenqiang and Li, Peihang and Qiu, Jiahao and Pei, Siqi and Ma, Lei and Huang, Tiejun and Wang, Mengdi and others},
  journal={arXiv preprint arXiv:2512.05941},
  year={2025}
}

@article{chainofground,
  title={Chain-of-Ground: Improving GUI Grounding via Iterative Reasoning and Reference Feedback},
  author={Li, Aiden Yiliu and Yu, Bizhi and Lei, Daoan and Ren, Tianhe and Liu, Shilong},
  journal={arXiv preprint arXiv:2512.01979},
  year={2025}
}

@article{nguyen2024improved,
  title={Improved gui grounding via iterative narrowing},
  author={Nguyen, Anthony},
  journal={arXiv preprint arXiv:2411.13591},
  year={2024}
}

@article{guispotlight,
  title={GUI-Spotlight: Adaptive Iterative Focus Refinement for Enhanced GUI Visual Grounding},
  author={Lei, Bin and Xu, Nuo and Payani, Ali and Hong, Mingyi and Liao, Chunhua and Cao, Yu and Ding, Caiwen},
  journal={arXiv preprint arXiv:2510.04039},
  year={2025}
}

@inproceedings{regionfocus,
  title={Visual test-time scaling for gui agent grounding},
  author={Luo, Tiange and Logeswaran, Lajanugen and Johnson, Justin and Lee, Honglak},
  booktitle={Proceedings of the IEEE/CVF International Conference on Computer Vision},
  pages={19989--19998},
  year={2025}
}

@article{uizoomer,
  title={UI-Zoomer: Uncertainty-Driven Adaptive Zoom-In for GUI Grounding},
  author={Tang, Fei and Chen, Bofan and Lu, Zhengxi and Chen, Tongbo and Nong, Songqin and Jiang, Tao and Xu, Wenhao and Lu, Weiming and Xiao, Jun and Zhuang, Yueting and others},
  journal={arXiv preprint arXiv:2604.14113},
  year={2026}
}

@inproceedings{dimogui,
  title={Dimo-gui: Advancing test-time scaling in gui grounding via modality-aware visual reasoning},
  author={Wu, Hang and Chen, Hongkai and Cai, Yujun and Liu, Chang and Ye, Qingwen and Yang, Ming-Hsuan and Wang, Yiwei},
  booktitle={Proceedings of the 2025 Conference on Empirical Methods in Natural Language Processing},
  pages={26257--26267},
  year={2025}
}

@article{reguide,
  title={ReGUIDE: Data Efficient GUI Grounding via Spatial Reasoning and Search},
  author={Lee, Hyunseok and Kim, Jeonghoon and Kim, Beomjun and Tack, Jihoon and Jo, Chansong and Lee, Jaehong and Park, Cheonbok and In, Sookyo and Shin, Jinwoo and Yoo, Kang Min},
  journal={arXiv preprint arXiv:2505.15259},
  year={2025}
}

@inproceedings{guirc,
  title={Test-time reinforcement learning for gui grounding via region consistency},
  author={Du, Yong and Yan, Yuchen and Tang, Fei and Lu, Zhengxi and Zong, Chang and Lu, Weiming and Jiang, Shengpei and Shen, Yongliang},
  booktitle={Proceedings of the AAAI Conference on Artificial Intelligence},
  volume={40},
  number={36},
  pages={30593--30601},
  year={2026}
}

@inproceedings{sspro,
  title={Screenspot-pro: Gui grounding for professional high-resolution computer use},
  author={Li, Kaixin and Meng, Ziyang and Lin, Hongzhan and Luo, Ziyang and Tian, Yuchen and Ma, Jing and Huang, Zhiyong and Chua, Tat-Seng},
  booktitle={Proceedings of the 33rd ACM International Conference on Multimedia},
  pages={8778--8786},
  year={2025}
}

@article{uivision,
  title={Ui-vision: A desktop-centric gui benchmark for visual perception and interaction},
  author={Nayak, Shravan and Jian, Xiangru and Lin, Kevin Qinghong and Rodriguez, Juan A and Kalsi, Montek and Awal, Rabiul and Chapados, Nicolas and {\"O}zsu, M Tamer and Agrawal, Aishwarya and Vazquez, David and others},
  journal={arXiv preprint arXiv:2503.15661},
  year={2025}
}

@article{mmbench,
  title={Mmbench-gui: Hierarchical multi-platform evaluation framework for gui agents},
  author={Wang, Xuehui and Wu, Zhenyu and Xie, JingJing and Ding, Zichen and Yang, Bowen and Li, Zehao and Liu, Zhaoyang and Li, Qingyun and Dong, Xuan and Chen, Zhe and others},
  journal={arXiv preprint arXiv:2507.19478},
  year={2025}
}

@inproceedings{osworldg,
  title={Scaling Computer-Use Grounding via User Interface Decomposition and Synthesis},
  author={Xie, Tianbao and Deng, Jiaqi and Li, Xiaochuan and Yang, Junlin and Wu, Haoyuan and Chen, Jixuan and Hu, Wenjing and Wang, Xinyuan and Xu, Yuhui and Wang, Zekun and others},
  booktitle={The Thirty-ninth Annual Conference on Neural Information Processing Systems Datasets and Benchmarks Track},
  year={2025}
}

@inproceedings{snell2025scaling,
  title={Scaling {LLM} Test-Time Compute Optimally Can be More Effective than Scaling Parameters for Reasoning},
  author={Charlie Victor Snell and Jaehoon Lee and Kelvin Xu and Aviral Kumar},
  booktitle={The Thirteenth International Conference on Learning Representations},
  year={2025},
  url={https://openreview.net/forum?id=4FWAwZtd2n}
}

%%%%%%%%%%%%%%%%%%%%%%%%%%%%%%%%%%%%%%%%%%%%%%%%%%%%%%%%%%%%

\newpage
\appendix

\section{Broader Impact}
\label{app:broader_impact}
This work proposes a training-free method for improving GUI grounding in vision-language models. By improving grounding accuracy without additional training, the method may benefit applications such as desktop automation, accessibility tools, and resource-efficient GUI agents. In addition, our analysis of prefill-stage behavior may encourage further research on more interpretable and reliable multimodal inference.
At the same time, improved GUI grounding may increase the capability of autonomous agents to interact with software environments, which could potentially be misused for unintended or unauthorized automation.

\section{Related Work}
\label{sec:relatedwork}

\subsection{GUI Grounding}

GUI grounding is a core capability for autonomous GUI agents~\citep{survey_nguyen2025gui, survey_zhang2024large, survey_wang2024gui, survey_tang2025survey}.
Early approaches address this task through supervised fine-tuning on grounding-specific datasets~\citep{showui, uground, cogagent}.
SeeClick~\citep{seeclick} formulates GUI grounding as a standalone pre-training objective and demonstrates that task-specific training improves agent performance.
Subsequent work extends this direction by introducing more diverse instruction types and cross-platform training data~\citep{osatlas, uground, ariaui, evocua, opencua}.
More recent studies apply reinforcement learning to GUI grounding~\citep{guir1, yuan2025enhancing, uivenus, guieyes}. These methods optimize task-level metrics and learn from interaction feedback.
GUI-Cursor~\citep{guicursor} models grounding as an iterative cursor movement process, where predictions are refined through visual feedback.
Other approaches improve grounding through multi-perspective instruction reasoning~\citep{uiins}, stabilized reward optimization~\citep{guirlvg}, and joint planning with grounding in agent frameworks~\citep{gta1}.
Despite strong performance, both supervised and reinforcement learning approaches require substantial training data and computational resources.

\subsection{Inference-Time Enhancement for GUI Grounding}
Inference-time scaling improves performance without additional training~\citep{snell2025scaling}.
In GUI grounding, one line of work focuses on zoom-in-based inference. These methods iteratively crop and zoom into candidate regions, increasing resolution and reducing the search space~\citep{zoomclick, chainofground, regionfocus, guispotlight, nguyen2024improved}. For example, ZoomClick~\citep{zoomclick} identifies the essential characteristics of zooming and enhances target-focused resolution through a repeated crop-and-zoom strategy. RegionFocus~\citep{regionfocus} apply dynamic zoom-in stragtegy to reduce background interference, thus enhancing grounding accuracy.
Another line improves grounding accuracy by aggregating predictions across multiple runs~\citep{mvp, uizoomer, dimogui, guirc, reguide}.
For instance, MVP~\citep{mvp} aggregates predictions from multiple cropped views to reduce variance. DiMo-GUI~\citep{dimogui} reasons over textual and iconic elements separately and selects the best candidate based on global analysis. UI-Zoomer~\citep{uizoomer} samples multiple candidate predictions and uses consensus voting together with adaptive cropping to achieve more robust grounding.
Despite their differences, these approaches share a common limitation. They do not modify the internal forward process of the model. Instead, they vary the input through iterative zoom-and-crop operations or aggregate outputs from multiple inferences.
As a result, each forward pass still contains a single prefill stage, and target selection must be completed in one attempt.

\section{Additional Qualitative Analysis}
\label{app:attn_visualization}
This appendix presents qualitative visualizations of query-position attention over visual tokens. For each example, we compare the baseline transition from prefill to the first decoding step with the corresponding transition under Re-Prefill, where a second prefill is applied before decoding. All results are produced using Qwen3-VL-8B-Instruct on ScreenSpot-Pro. Ground-truth targets are marked with blue rectangles, and predicted coordinates are shown as orange hollow circles for visibility.

\noindent{\textbf{Prefill vs. Re-Prefill. }}
Under the baseline, attention at the prefill stage is distributed across multiple candidate regions, consistent with the high variance observed in Figure~\ref{fig:analysis}(b). While these regions often include the ground-truth target, distractors with similar appearance or semantics are also activated, making disambiguation difficult in a single forward pass.
With Re-Prefill, attention becomes more concentrated on the correct region, while responses on distractor regions are noticeably suppressed. As a result, the effective candidate set is reduced, leading to more reliable target selection before decoding. This supports the design of Re-Prefill, where key visual tokens extracted from the first prefill guide a second prefill to refine attention within candidate regions and improve target selection.

\noindent{\textbf{The first decoding step. }}
At the first decoding step, attention contracts sharply to a localized region in both settings, consistent with the drop in spatial variance from $t=0$ to $t=1$ in Figure~\ref{fig:analysis}(b). The predicted coordinate is placed within this region, indicating that target selection has already been determined before decoding.
With Re-Prefill, this localized region is more likely to align with the ground-truth target, whereas the baseline may converge to a distractor region. This demonstrates that errors introduced during prefill persist into decoding, while improvements from Re-Prefill are directly propagated to coordinate prediction, consistent with the prefill bottleneck identified in Section~\ref{sec:bottleneck}.

\begin{figure}[t]
  \centering
  \includegraphics[width=\textwidth]{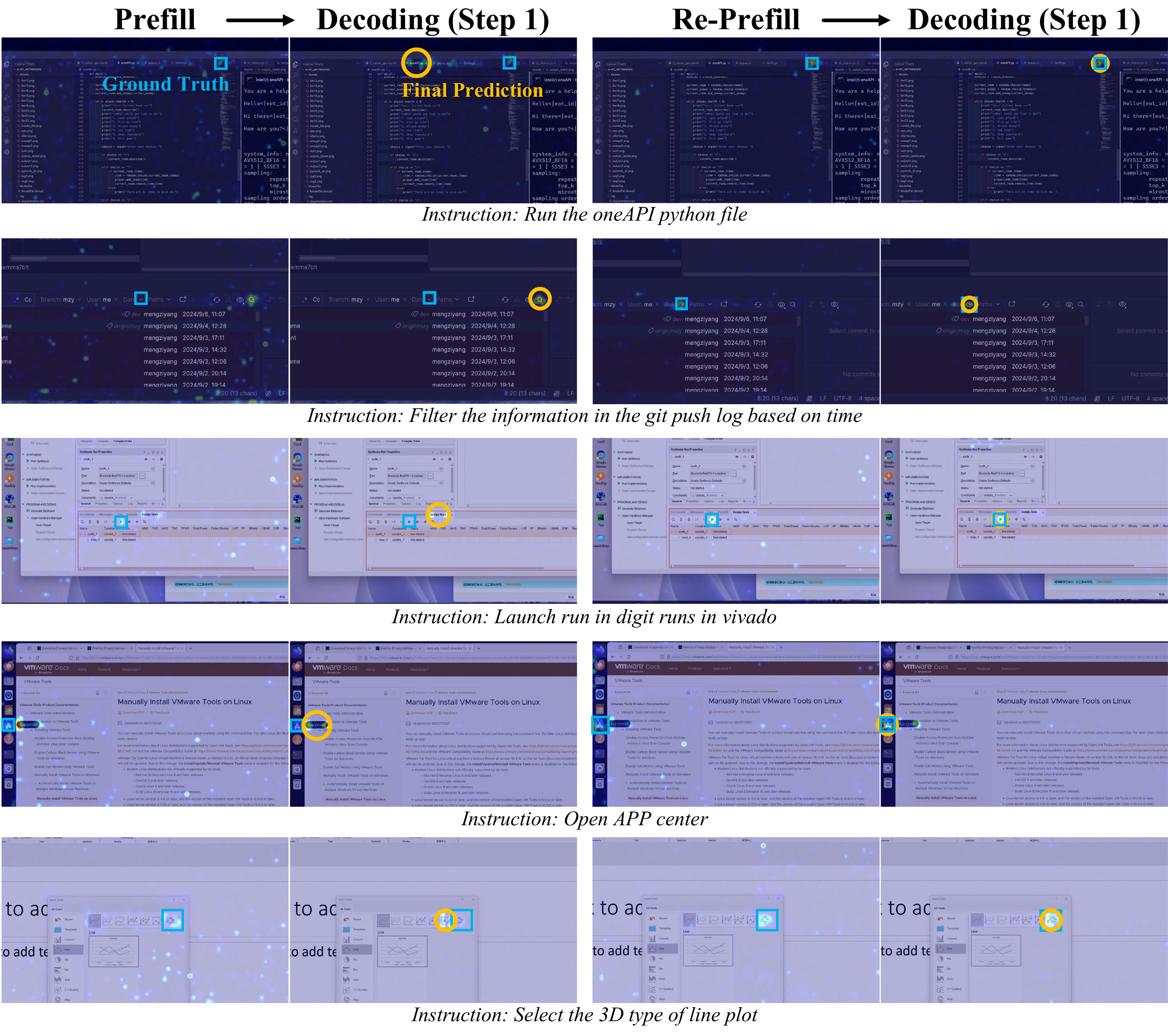}
  \caption{
    \textbf{Query-position attention heatmaps across stages on ScreenSpot-Pro.}
    Each row shows one example. Columns 1–2 present the baseline transition from prefill to the first decoding step, while Columns 3–4 show the corresponding transition with Re-Prefill. The blue rectangle marks the ground-truth target, and the orange circle indicates the predicted coordinate.
    Compared to the baseline, Re-Prefill focuses attention on the correct region during prefill, suppresses distractors, and improves localization for subsequent decoding.
  }
  \label{fig:attn_visualization}
  \vspace{-1.0em}
\end{figure}

\section{Additional Analysis of Prefill-Stage Attention Dynamics}
\label{app:cross_bench}

\begin{figure}[t]
  \centering
  \includegraphics[width=\textwidth]{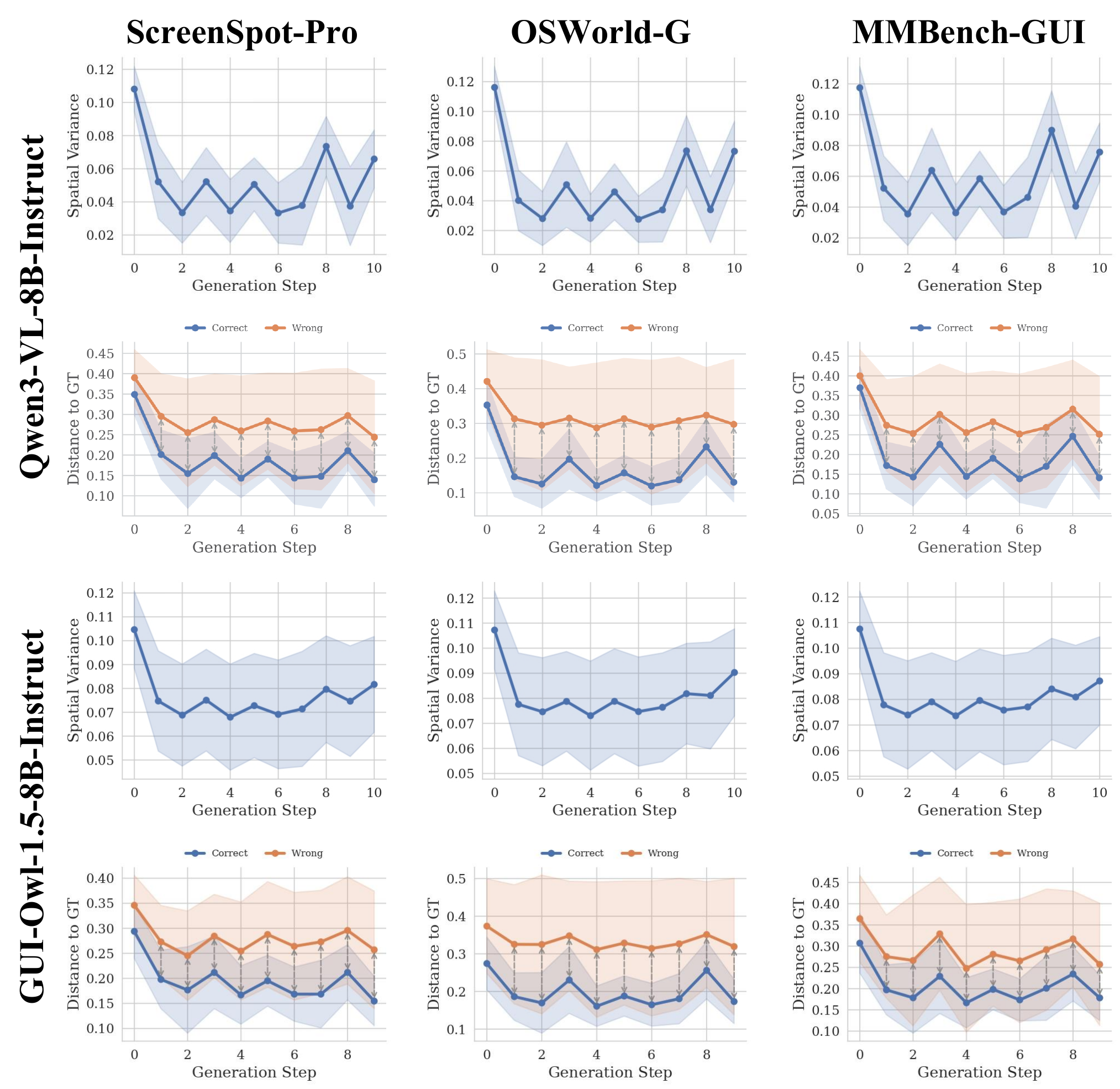}
  \caption{
    \textbf{Spatial variance and prefill-stage error analysis across models and benchmarks.}
    Rows 1--2 show results for Qwen3-VL-8B-Instruct, and Rows 3--4 for GUI-Owl-1.5-8B-Instruct.
    For each model, the first row shows spatial variance across generation steps, and the second row shows attention-centroid deviation for correct and incorrect predictions.
    Across all settings, attention is dispersed at prefill and rapidly contracts after the first decoding step, while errors introduced at prefill persist throughout decoding.
  }
  \label{fig:cross_benchmark}
  \vspace{-1.0em}
\end{figure}

This appendix extends the analysis in Section~\ref{sec:bottleneck} to multiple models and benchmarks.
Experiments are conducted on two VLMs, Qwen3-VL-8B-Instruct and GUI-Owl-1.5-8B-Instruct, across three datasets, ScreenSpot-Pro, OSWorld-G, and MMBench-GUI.
Figure~\ref{fig:cross_benchmark} summarizes the results.
The first two rows correspond to Qwen3-VL-8B-Instruct, and the last two rows correspond to GUI-Owl-1.5-8B-Instruct. The columns correspond to ScreenSpot-Pro, OSWorld-G, and MMBench-GUI.
For each model, the first row shows spatial variance across generation steps, and the second row shows attention-centroid deviation from the ground truth for correct and incorrect predictions.

Consistent patterns are observed across all models and datasets.
Spatial variance is high at the prefill stage ($t=0$), drops sharply after the first decoding step ($t=1$), and remains low thereafter, indicating rapid concentration of attention to a single region.
In parallel, incorrect predictions are already misaligned with the ground truth at $t=0$ and remain so during decoding, whereas correct predictions stay well aligned.
These results show that the two-stage inference behavior identified in Section~\ref{sec:bottleneck} holds across model architectures and benchmark domains, indicating that the prefill-stage bottleneck is a general property of VLM-based GUI grounding rather than an artifact of a specific model or dataset.

\begin{table}[t]
\centering
\caption{Results on OSWorld-G (refined annotation). Methods marked with \repro{} are reproduced using the zoom-in strategy adopted in this work.}
\label{tab:main_osworld_refined}
\resizebox{\columnwidth}{!}{%
\begin{tabular}{l c c c c c >{\columncolor{blue!10}}c}
\toprule
\textbf{Model} & \textbf{Text Match.} & \textbf{Element Rec.} & \textbf{Layout Under.} & \textbf{Fine-grained Manip.} & \textbf{Refusal} & \multicolumn{1}{c}{\cellcolor{white}\textbf{Overall}} \\
\midrule
UI-TARS-1.5-7B~\citep{uitars15} & 52.6 & 75.4 & 72.4 & 66.7 & - & 64.2 \\
% GTA1-7B~\citep{gta1} & 63.2 & 82.1 & 74.2 & 70.5 & - & 67.7 \\
GTA1-32B~\citep{gta1} & 63.2 & 83.6 & 84.4 & 70.5 & - & 72.2 \\
% OpenCUA-32B~\citep{opencua} & 63.2 & 79.9 & 84.9 & 62.1 & 7.4 & 70.2 \\
MAI-UI-8B~\citep{maiui} & 79.3 & 78.8 & 84.2 & 59.7 & - & 72.9 \\
GUI-Owl-1.5-8B-Instruct~\citep{guiowl} & 73.1 & 69.4 & 71.0 & 74.8 & 42.5 & 69.3 \\
\midrule
\multicolumn{7}{l}{\textit{Training-free methods}} \\
MVP-8B~\citep{mvp} & 82.4 & 76.7 & 79.5 & 61.7 & - & 72.7 \\
MVP-32B~\citep{mvp} & 81.6 & 76.4 & 77.5 & 62.4 & - & 72.0 \\
\midrule

\textit{\textcolor{gray}{Qwen3-VL-8B-Instruct~\citep{qwen3vl}}}\repro &
\textcolor{gray}{78.9} & \textcolor{gray}{78.2} & \textcolor{gray}{80.2} & \textcolor{gray}{57.7} & \textcolor{gray}{-} & \textcolor{gray}{71.9} \\
\ \ + Re-Prefill & 81.2 & 79.1 & 82.2 & 61.7 & - & 73.8\up{1.9} \\
\midrule

\textit{\textcolor{gray}{MAI-UI-8B~\citep{maiui}}}\repro &
\textcolor{gray}{79.7} & \textcolor{gray}{80.0} & \textcolor{gray}{81.8} & \textcolor{gray}{58.4} & \textcolor{gray}{-} & \textcolor{gray}{73.2} \\
\ \ + Re-Prefill & 81.2 & 79.7 & 81.8 & 59.1 & - & 73.5\up{0.3} \\
\midrule

\textit{\textcolor{gray}{GUI-Owl-1.5-8B-Instruct~\citep{guiowl}}}\repro &
\textcolor{gray}{81.6} & \textcolor{gray}{79.4} & \textcolor{gray}{82.6} & \textcolor{gray}{65.1} & \textcolor{gray}{-} & \textcolor{gray}{74.6} \\
\ \ + Re-Prefill & 81.6 & 79.7 & 83.4 & 65.8 & - & 75.0\up{0.4} \\
\midrule

\textit{\textcolor{gray}{Qwen3-VL-32B-Instruct~\citep{qwen3vl}}}\repro &
\textcolor{gray}{82.4} & \textcolor{gray}{85.2} & \textcolor{gray}{87.0} & \textcolor{gray}{62.4} & \textcolor{gray}{-} & \textcolor{gray}{77.3} \\
\ \ + Re-Prefill & 83.5 & 84.2 & 85.8 & 65.8 & - & 77.5\up{0.2} \\
\bottomrule
\end{tabular}%
}
\end{table}

\section{Results on OSWorld-G-Refined}
\label{app:osworldg_refined}

This appendix presents results on OSWorld-G with the refined annotation.
As shown in Table~\ref{tab:main_osworld_refined}, Re-Prefill improves all base models under the refined annotation, with gains of $+1.9$ on Qwen3-VL-8B-Instruct, $+0.3$ on MAI-UI-8B, $+0.4$ on GUI-Owl-1.5-8B-Instruct, and $+0.2$ on Qwen3-VL-32B-Instruct.
Compared with MVP using the same base models, Re-Prefill reaches $73.8$ versus $72.7$ for MVP-8B at the 8B scale and $77.5$ versus $72.0$ for MVP-32B at the 32B scale.
These results demonstrate that the effectiveness of Re-Prefill remains consistent under refined annotations.

\section{Computational Efficiency Analysis}
\label{app:efficiency}
We compare the computational efficiency of Re-Prefill with existing training-free methods, including MVP~\citep{mvp} and ZoomClick~\citep{zoomclick}, under identical base models.
Experiments are conducted on Qwen3-VL-8B-Instruct and Qwen3-VL-32B-Instruct across three benchmarks, ScreenSpot-Pro, OSWorld-G, and MMBench-GUI.
We report the average inference time per sample (in seconds) and the corresponding grounding accuracy. To ensure fairness, MVP and ZoomClick are reproduced under the same experimental environment and settings.

As shown in Table~\ref{tab:comp_overhead}, Re-Prefill achieves the highest accuracy among training-free methods across all datasets and model scales while maintaining competitive runtime.
On 8B models, Re-Prefill is slightly faster than or comparable to existing methods. 
On 32B models, Re-Prefill operates under a comparable computational budget to other training-free methods while achieving higher grounding performance.
% This difference is mainly due to the additional prefill stage and the enhanced visual representations introduced during decoding, both of which operate on higher-dimensional hidden states in larger models.
% In contrast, ZoomClick shows higher latency on high-resolution datasets such as ScreenSpot-Pro, as it dynamically performs multiple cropping operations based on image resolution.
Overall, Re-Prefill provides a favorable efficiency–performance trade-off.
It improves grounding accuracy with competitive efficiency, particularly on smaller models.

\begin{table}[t]
\caption{Comparison of computational efficiency among training-free methods. Inference time is reported as the average per-sample latency in seconds. Methods marked with \repro{} are reproduced under the same experimental settings in this work.}
\label{tab:comp_overhead}
\footnotesize
\centering
\begin{tabular}{l cc cc cc}
\toprule
\makecell[l]{\\ \textbf{Methods} \\}
& \multicolumn{2}{c}{\textbf{ScreenSpotPro}}\vspace{-1pt} 
& \multicolumn{2}{c}{\textbf{OSWorld-G}}\vspace{-1pt}  
& \multicolumn{2}{c}{\textbf{MMBench-GUI}}\vspace{-1pt} \\
\cmidrule(lr){2-3} \cmidrule(lr){4-5} \cmidrule(lr){6-7}
 & Time (s)$\downarrow$ & Acc. (\%)$\uparrow$ 
 & Time (s)$\downarrow$ & Acc. (\%)$\uparrow$ 
 & Time (s)$\downarrow$ & Acc. (\%)$\uparrow$ \\
\midrule

\multicolumn{7}{l}{\textit{Qwen3-VL-8B-Instruct}} \vspace{2pt} \\
MVP\repro  & 9.0 & 65.7 & 4.2 & 59.3 & 5.4 & 84.4 \\
ZoomClick\repro  & 11.2 & 66.1 & \textbf{4.1} & 56.7 & 7.0 & 84.4 \\
Re-Prefill & \textbf{8.0} & \textbf{70.1} & \textbf{4.1} & \textbf{65.7} & \textbf{4.8} & \textbf{86.4} \\

\midrule

\multicolumn{7}{l}{\textit{Qwen3-VL-32B-Instruct}} \vspace{2pt} \\
MVP\repro        & \textbf{11.8} & 73.9 & 6.4 & 65.3 & 8.4 & 87.8 \\
ZoomClick\repro  & 12.4 & 71.5 & \textbf{6.2} & 58.2 & \textbf{7.6} & 86.4 \\
Re-Prefill & 12.2 & \textbf{76.8} & 7.5 & \textbf{70.1} & 7.8 & \textbf{89.5} \\

\bottomrule
\end{tabular}
\end{table}

\section{Pseudocode of Re-Prefill}
\label{app:algorithm}

Algorithm~\ref{alg:reprefill} gives the complete procedure of Re-Prefill described in Section~\ref{sec:reprefill}.
Given an input sequence $\mathbf{x} = [\mathbf{S}; \mathbf{V}; \mathbf{T}]$, the algorithm performs an initial prefill, 
selects key visual tokens through cross-layer consistency, executes a layer-wise second prefill with hyperparameter $L_c$, 
and composes the decoding context $\mathcal{C}^{*}$.

\begin{algorithm}[!htbp]
\caption{Re-Prefill Inference}
\label{alg:reprefill}
\begin{algorithmic}[1]
\REQUIRE Input sequence $\mathbf{x} = [\mathbf{S}; \mathbf{V}; \mathbf{T}]$; quantile threshold $\rho$; ratio threshold $\gamma$; continuity layers $L_c$
\ENSURE Predicted coordinates $(x, y)$
\vspace{0.5em}
\STATE \textbf{Step 1: Initial prefill and attention extraction}
\STATE $\bigl(\mathbf{h}_q,\, \mathcal{C}_1,\, \{\mathbf{a}^{(l)}\}_{l=1}^{L},\, [\tilde{\mathbf{S}};\tilde{\mathbf{V}};\tilde{\mathbf{T}}]\bigr) \leftarrow f_{\mathrm{prefill}}(\mathbf{x})$
\vspace{0.8em}
\STATE \textbf{Step 2: Key visual token selection}
\STATE $\tau \leftarrow \mathrm{quantile}\bigl(\{\mathbf{a}^{(l)}\}_{l=1}^{L},\; \rho\bigr)$
\FOR{each visual token index $i = 1,\dots,N_v$}
    \STATE $r(i) \leftarrow \frac{1}{L} \sum_{l=1}^{L} \mathds{1}[\,a^{(l)}(i) > \tau\,]$
\ENDFOR
\STATE $\tilde{\mathbf{V}}^{*} \leftarrow \{\, \tilde{v}_i \mid r(i) \geq \gamma \,\}$
\vspace{0.8em}
\STATE \textbf{Step 3: Layer-wise second prefill}
\FOR{decoder layer $l = 1,\dots,L$}
    \IF{$l \le L_c$}
        \STATE $\mathrm{prefix} \leftarrow [\,\tilde{\mathbf{S}};\,\tilde{\mathbf{V}};\,\tilde{\mathbf{T}}\,]$
    \ELSE
        \STATE $\mathrm{prefix} \leftarrow [\,\tilde{\mathbf{V}}^{*};\,\tilde{\mathbf{T}}\,]$
    \ENDIF
    \STATE apply decoder layer $l$ to $[\,\mathrm{prefix};\,\mathbf{S};\,\mathbf{V};\,\mathbf{T}\,]$
\ENDFOR
\STATE Extract enriched visual representations $\hat{\mathbf{V}}$ at the second set of visual-token positions
\vspace{0.8em}
\STATE \textbf{Step 4: Decoding context composition and generation}
\STATE $\mathcal{C}^{*} \leftarrow [\,\tilde{\mathbf{S}};\,\tilde{\mathbf{V}};\,\tilde{\mathbf{T}};\,\hat{\mathbf{V}}\,]$
\STATE $(x, y) \leftarrow f_{\mathrm{decode}}(\mathcal{C}^{*})$
\RETURN $(x, y)$
\end{algorithmic}
\end{algorithm}
% \vspace{-1.0em}

\section{More Discussions}
\label{app:more_discussions}

This section addresses three concerns regarding the motivation and design of Re-Prefill that are not fully unpacked in the main text.

\noindent{\textit{\textbf{Q1. Are spatial variance and centroid sufficient indicators of target selection?}}}

Spatial variance and centroid characterize the attention distribution rather than directly measuring target selection. In particular, high attention does not strictly imply the correct target, as unrelated regions may also receive strong responses.
This limitation is explicitly accounted for in our design. Instead of treating attention as a precise indicator, we use it as a coarse, recall-oriented signal to identify candidate regions. The key visual token selection step aims to ensure that the correct target region is included, even in the presence of noisy attention. The actual target selection refinement is then performed by the second prefill under this candidate prior.
Under this design, spatial variance and centroid are not used as direct measures of target selection, but to reveal the existence and evolution of a target-selection signal during prefill. 
% This also explains the insensitivity to $\rho$ and $\gamma$. As long as the selected set sufficiently covers candidate regions, the second prefill can refine the final prediction regardless of the exact threshold values.

\noindent{\textit{\textbf{Q2. Why does Re-Prefill not amplify first-prefill errors?}}}

Although Re-Prefill relies on first-prefill attention to select key visual tokens, it does not amplify errors because the initial attention rarely concentrates exclusively on a wrong region. Instead, it spreads over multiple candidate UI regions, typically including the ground-truth target with non-trivial attention mass, as shown in Figure~\ref{fig:analysis}(a) and Appendix~\ref{app:attn_visualization}. 
Even for incorrect predictions, the centroid deviation remains bounded (Figure~\ref{fig:analysis}(c)), indicating that the correct region is not excluded.
Thus, the difference between correct and incorrect cases lies in how attention is distributed among candidates rather than whether the target is covered. The second prefill re-evaluates these candidates under an explicit prior formed by key visual tokens. As long as the correct target is included, it can be promoted during re-ranking, enabling error correction rather than amplification.

\noindent{\textit{\textbf{Q3. Why are the performance gaps to Blind Re-Prefill and Random Token Selection in Table~\ref{tab:ablation_component} relatively small?}}}

The relatively small gaps to Blind Re-Prefill ($0.7\%$) and Random Token Selection ($1.0\%$), compared to the $4.1\%$ gap to Embedding Addition, reflect different components of the method rather than comparable variants. The large gap to Embedding Addition isolates the contribution of the second prefill, showing that re-encoding visual tokens is the primary source of improvement.
Given the second prefill, the remaining gains ($0.7\text{--}1.0\%$) come from token selection. Blind Re-Prefill dilutes the signal by treating all visual tokens equally, while Random Selection introduces misaligned guidance. In contrast, selecting key visual tokens concentrates attention on candidate regions and aligns the guidance signal. Overall, the second prefill provides the main gain, and selection improves its effectiveness.

\section{Limitations}
\label{app:limitations}

While Re-Prefill demonstrates consistent improvements across models and benchmarks, several limitations remain.
% \noindent{\textbf{Additional computational overhead.}}
% Re-Prefill introduces a second prefill stage, which increases inference cost compared to standard single-pass decoding. Although the method remains training-free and does not require multiple full forward passes as in iterative cropping or ensembling approaches, the added computation may impact latency in real-time applications.
% \noindent{\textbf{Heuristic token selection.}}
% The cross-layer consistency filter for selecting key visual tokens relies on heuristic criteria, including a global attention threshold and a layer-wise consistency ratio. Although effective in practice, these choices are not guaranteed to be optimal across models or tasks. Future work may investigate more principled or learned selection strategies that better exploit the information encoded in model representations.
% \noindent{\textbf{Limited gains in instruction-dominant scenarios.}}
As shown in the experimental results, the improvements from Re-Prefill are more pronounced in visually complex settings with high resolution and dense UI elements, where accurate target selection is the main challenge. In contrast, gains are smaller in instruction-heavy settings, where performance is more constrained by the ability to interpret complex or abstract instructions. This suggests that while Re-Prefill improves target selection during the prefill stage, it does not directly enhance instruction understanding. Extending the approach to better support complex instruction comprehension remains an important direction for future work.
%%%%%%%%%%%%%%%%%%%%%%%%%%%%%%%%%%%%%%%%%%%%%%%%%%%%%%%%%%%%

% \newpage
% \input{checklist.tex}

\end{document}